\newif\ifnbld
\newif\ifconf
\newif\iftr
\newif\ifEXT
\EXTfalse

\trtrue
\conffalse

\nbldtrue

\documentclass{article}

\usepackage{microtype}
\usepackage{graphicx}
\usepackage{subcaption}
\usepackage{booktabs} 

\ifconf
\usepackage{icml2026}
\fi

\iftr
\usepackage[preprint]{icml2026_custom}
\fi

\usepackage[hidelinks]{hyperref}

\usepackage{amsmath}
\usepackage{amssymb}
\usepackage{mathtools}
\usepackage{amsthm}

\usepackage[capitalize,noabbrev]{cleveref}

\theoremstyle{plain}

\theoremstyle{definition}

\theoremstyle{remark}

\usepackage[textsize=tiny]{todonotes}

\usepackage[most]{tcolorbox} 
\usepackage{xurl} 

\newcommand{\name}{Multi-Head RAG}
\newcommand{\nameS}{Multi-Head RAG\ }
\newcommand{\nameA}{MRAG}
\newcommand{\nameAS}{MRAG\ }

\newif\ifsq     

\newif\ifsqCAP
\newif\ifsqVS
\newif\ifsqEN
\newif\ifsqTIT

\newcommand{\ignore}[1]{}

\sqtrue
\sqCAPtrue
\sqENtrue
\sqVStrue
\sqTITtrue

\iftr
\sqfalse
\fi
\sqCAPfalse

\usepackage{balance}
\usepackage{epstopdf}
\usepackage{placeins}

\usepackage{graphicx}
\usepackage{float}
\usepackage{multirow}
\usepackage{rotating}
\usepackage{makecell}
\usepackage{tabulary}
\usepackage{parcolumns}
\usepackage{tikz}
\usetikzlibrary{tikzmark}

\usepackage{xpatch}
\expandafter\xpatchcmd
\csname pgfk@/tikz/every picture/.@cmd\endcsname
{\thepage}{\arabic{page}}{}{}

\tikzstyle{comment} = [draw, fill=blue!70, text=white, text width=3cm, minimum height=1cm, rounded corners, align=left, font=\scriptsize]
\tikzstyle{background_alg} = [draw, fill=blue!20, opacity=0.4, inner sep=4pt, rounded corners=2pt]

\usetikzlibrary{shapes}
\usetikzlibrary{plotmarks}
\usetikzlibrary{calc, fit}

\usepackage{enumitem}

\usepackage{amsthm}
\usepackage{amsmath,amssymb,amsfonts}
\usepackage{mathtools,mathrsfs}

\usepackage{soul}
\usepackage{fontawesome}
\usepackage{pifont}
\usepackage{textcomp}
\usepackage{booktabs}
\usepackage{url}
\usepackage{pbox}
\usepackage[normalem]{ulem}
\usepackage[10pt]{moresize}

\newcommand{\noAnswer}{\textcolor{black}{\faQuestionCircle}}

\ifsqCAP
\usepackage[font={normalfont, footnotesize}]{caption}
\usepackage[font={normalfont, footnotesize}]{subcaption}
\else
\usepackage[font={normalfont, footnotesize}]{caption}
\usepackage[font={normalfont, footnotesize}]{subcaption}
\fi

\newcommand{\vspaceSQ}[1]{\ifsqVS\vspace{#1}\fi}
\newcommand{\enlargeSQ}[1]{\ifsqEN\enlargethispage{\baselineskip}\fi}

\ifsqTIT
\usepackage[compact]{titlesec}
\titlespacing*{\section}{0pt}{0pt}{-1pt}
\titlespacing*{\subsection}{0pt}{0pt}{-3pt}
\titlespacing*{\subsubsection}{0pt}{2pt}{1pt}
\fi

\usepackage{xcolor}
\definecolor{darkgrey}{RGB}{70,70,70}
\definecolor{lightgrey}{RGB}{200,200,200}
\definecolor{lyellow}{RGB}{255,255,100}
\definecolor{llyellow}{RGB}{250,250,180}
\definecolor{lgreen}{RGB}{144,238,144}
\definecolor{raphael_comments}{RGB}{13, 145, 24}

\usepackage[customcolors]{hf-tikz}
\hfsetbordercolor{white}
\hfsetfillcolor{vlgray}

\definecolor{vlgray}{rgb}{0.77 0.77 0.77}
\definecolor{ablack}{rgb}{0.2 0.2 0.2}
\definecolor{vllgray}{rgb}{0.9 0.9 0.9}
\definecolor{bblue}{rgb}{0.7 0.7 0.99}

\usepackage{colortbl}

\usepackage{inconsolata}
\usepackage{listings}

\ifsq
\else
\fi

\newcommand{\maciej}[1]{\textcolor{blue}{[Maciej: #1]}}

\newcommand{\ales}[1]{\textcolor{blue}{[Ales: #1]}}

\newcommand{\robert}[1]{\textcolor{blue}{[Robert: #1]}}

\definecolor{hlL}{rgb}{0.8 0.8 0.99}

\newcounter{highlight}

\newcounter{hlLR}

\newcounter{hlLIR}

\newcounter{hlLIIR}

\newcounter{Ahighlight}

\renewcommand{\epsilon}{\ensuremath\varepsilon}

\renewcommand{\phi}{\ensuremath{\varphi}}

\definecolor{c_green}{HTML}{339933}
\definecolor{c_blue}{HTML}{333399}
\definecolor{c_red}{HTML}{993333}
\definecolor{c_gray}{HTML}{333333}
\definecolor{c_cyan}{HTML}{339999}
\definecolor{c_magenta}{HTML}{993399}
\definecolor{c_yellow}{HTML}{999933}

\newcommand{\istar}{\raisebox{-0.125em}{\includegraphics[height=0.75em]{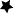}}}
\newcommand{\iparta}{\raisebox{-0.125em}{\includegraphics[height=0.75em]{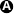}}}
\newcommand{\ipartb}{\raisebox{-0.125em}{\includegraphics[height=0.75em]{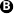}}}
\newcommand{\ipartc}{\raisebox{-0.125em}{\includegraphics[height=0.75em]{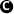}}}
\newcommand{\ipartd}{\raisebox{-0.125em}{\includegraphics[height=0.75em]{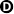}}}
\newcommand{\idoc}{\raisebox{-0.125em}{\includegraphics[height=0.75em]{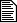}}}
\newcommand{\ichunk}{\raisebox{0em}{\includegraphics[height=0.5em]{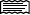}}}
\newcommand{\iemb}{\raisebox{-0.125em}{\includegraphics[height=0.75em]{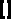}}}

\newcommand{\iquery}{\raisebox{-0.125em}{\includegraphics[height=0.75em]{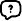}}}

\newcommand{\idmatch}{\raisebox{-0.125em}{\includegraphics[height=0.75em]{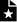}}}
\newcommand{\icmatch}{\raisebox{-0.125em}{\includegraphics[height=0.75em]{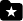}}}
\newcommand{\icrematch}{\raisebox{-0.125em}{\includegraphics[height=0.75em]{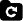}}}
\newcommand{\inomatch}{\raisebox{-0.125em}{\includegraphics[height=0.75em]{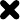}}}

\usepackage{algorithm}
\usepackage{algorithmic}
\usepackage{wrapfig}

\renewcommand{\maciej}[1]{}

\usepackage{stmaryrd}

\newcommand{\faY}[0]{\faBatteryFull}
\newcommand{\faH}[0]{\faBatteryHalf}
\newcommand{\faN}[0]{\faTimes}
\newcommand{\faU}[0]{\noAnswer}

\renewcommand{\arraystretch}{1.0}

\definecolor{lightgray1}{gray}{0.6}
\definecolor{lightgray2}{gray}{0.8}

\renewcommand{\marginpar}[1]{}

\usepackage{tabularx}

\newcommand{\mytitle}{\name: Solving Multi-Aspect Problems with LLMs}

\icmltitlerunning{\mytitle}

\begin{document}

\twocolumn[
  \icmltitle{\mytitle}



  \icmlsetsymbol{equal}{*}

  \begin{icmlauthorlist}
    \icmlauthor{Maciej Besta}{eth}
    \icmlauthor{Ales Kubicek}{eth}
    \icmlauthor{Robert Gerstenberger}{eth}
    \icmlauthor{Marcin Chrapek}{eth}
    \icmlauthor{Roman Niggli}{eth}
    \icmlauthor{Patrik Okanovic}{eth}
    \icmlauthor{Yi Zhu}{eth}
    \icmlauthor{Patrick Iff}{eth}
    \icmlauthor{Michal Podstawski}{nas}
    \icmlauthor{Lucas Weitzendorf}{eth}
    \icmlauthor{Mingyuan Chi}{eth}
    \icmlauthor{Joanna Gajda}{cle}
    \icmlauthor{Piotr Nyczyk}{ide,cle}
    \icmlauthor{J\"{u}rgen M\"{u}ller}{bas}
    \icmlauthor{Hubert Niewiadomski}{ide,cle}
    \icmlauthor{Torsten Hoefler}{eth}
  \end{icmlauthorlist}

  \icmlaffiliation{eth}{Department of Computer Science, ETH Zurich, Zurich, Switzerland}
  \icmlaffiliation{cle}{Cledar, Wieliczka, Poland}
  \icmlaffiliation{ide}{IDEAS Research Institute, Warsaw, Poland}
  \icmlaffiliation{nas}{NASK National Research Institute, Warsaw, Poland}
  \icmlaffiliation{bas}{BASF SE, Ludwigshafen, Germany}

  \icmlcorrespondingauthor{Maciej Besta}{maciej.besta@inf.ethz.ch}


  \vskip 0.3in
]



\printAffiliationsAndNotice{}  

\begin{abstract}
Retrieval-Augmented Generation (RAG) improves Large Language Models (LLMs) by retrieving supporting documents into the prompt, but existing methods do not explicitly target queries that require fetching multiple documents with \textit{substantially} different content. Such \textit{multi-aspect} queries are challenging because relevant documents can be far apart in embedding space, making joint retrieval difficult. We introduce \nameS\ (\nameA), which addresses this gap with a simple yet powerful idea: using Transformer multi-head attention activations rather than the standard decoder-layer embedding, as retrieval keys. It leverages the observation that different heads capture different semantic aspects. This yields multi-aspect embeddings for both documents and queries, improving retrieval accuracy on complex queries. We show \nameA's design advantages over 18 RAG baselines, up to 20\% higher retrieval success ratios for real-world use cases, and improved downstream LLM generation. \nameAS integrates seamlessly with existing RAG frameworks and benchmarks.
%
%
\end{abstract}

\ifnbld
\begin{center}
\textbf{Website \& code:} {\url{https://github.com/spcl/MRAG}}
\end{center}
\fi

\vspaceSQ{-0.5em}
\section{Introduction}
\label{sec:intro}

\iftr

Retrieval-Augmented Generation (RAG)~\citep{lewis2020retrieval, guu2020realm} emerged as a promising remedy for several key limitations of Large Language Models (LLMs). By decoupling knowledge from model weights, RAG reduces the risk of leaking confidential data~\citep{yan2024protecting, shi2024detecting, patil2024can}, a critical concern when training on sensitive corpora. It also mitigates hallucinations~\citep{zhang2023siren, xu2024hallucination, huang2023survey} by grounding LLM outputs in retrieved, verifiable information. The core mechanism involves augmenting a generative LLM with a retrieval module that fetches relevant passages from an external corpus in response to a query. Rather than relying solely on static, parametric knowledge, RAG dynamically incorporates retrieved content into the model's context, enabling more accurate and up-to-date responses. While early RAG systems required training task-specific retrievers and readers~\citep{humeau2019poly, singh2021end}, the current trend favors lightweight, in-context learning (ICL) approaches~\citep{gao2024retrieval, zhao2024retrieval, hu2024rag, huang2024survey}, which avoid the cost and complexity of retraining and allow for rapid knowledge updates without modifying the underlying LLM parameters.

A RAG pipeline consists of two main stages: data preparation and query execution. In the preparation stage, a vector database (DB) is constructed by embedding a collection of documents and storing these embeddings alongside their associated content. At inference time, the query is similarly embedded, and nearest-neighbor search retrieves the most relevant data items, which are then passed to the LLM for final answer generation. Ongoing developments in RAG have led to a variety of designs~\citep{gao2024retrieval, zhao2024retrieval, hu2024rag, huang2024survey, yu2024evaluation, mialon2023augmented, li2022survey, chen2024benchmarking, xiong2024benchmarking, lyu2024crud, es2023ragas, sarthi2024raptor, asai2023selfrag, abdallah2024generator, delile2024graph, edge2024from, manathunga2023retrieval, zeng2024federated, wewer2021updating, xu2024active, huang2025retrieval, yang2025superrag, su2025parametric, zhao2025moc, zhao2024meta, chen2024hiqa, chan2024rq, jiang2024ragraph, yu2023chain}.

\fi

Yet, no existing RAG method or benchmark explicitly targets \textit{multi-aspect} problems, that is, queries requiring the integration of multiple, semantically distinct aspects. For example, answering ``What car did Alexander the Great drive?'' (assuming no historical pretraining) requires retrieving unrelated documents on Alexander the Great and on car manufacturing, whose embeddings may lie far apart in the vector space.
\iftr
Such multi-aspect queries are common in industrial settings, as confirmed by extensive discussions with our industry collaborators and further supported by our analysis of over 35 industry reports~\citep{fra2003human, gehman2003columbia, zografos2013decision, reason2006revisiting, kalia2013personalized, packham2017columbia, messiou2023multimodal, multer2013developing, coury2010transportation, harle2017investigation, o2000wheel, gordon2005designing, bridger2021guide, opentext} (details are in Appendix~\ref{sec:appendix-aspects}; we considered accident prevention, healthcare, airport management, and other domains).
\fi
For example, in a chemical plant accident, determining the cause might require accessing diverse and confidential documents related to worker psychology (\textit{``Was it mismanagement?''}), equipment records (\textit{``Was a part outdated or rusty?''}), weather conditions (\textit{``Were there power spikes due to a storm?''}), or even microclimate (\textit{``Was prolonged humidity a factor?''}). As shown in Section~\ref{sec:eval}, such cases have been unaddressed by modern RAG schemes and benchmarks~\citep{chen2024benchmarking, xiong2024benchmarking, lyu2024crud, es2023ragas}.


\enlargeSQ

\setlength{\columnsep}{10pt}%
\begin{wrapfigure}{r}{0.5\columnwidth}
\centering
\vspace{-1em}
\includegraphics[width=0.5\columnwidth]{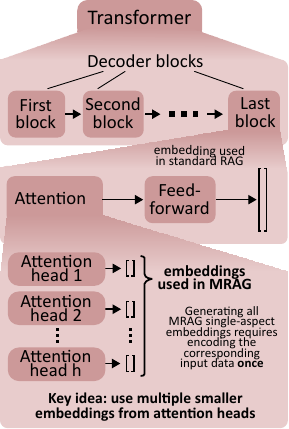}
\vspace{-1.5em}
\caption{A comparison of the generation of standard RAG and \nameS embeddings.}
\label{fig:embedding-model}
\vspaceSQ{-1.25em}
\end{wrapfigure}

In this work, we propose \nameS\ (\nameA): a scheme that addresses the above problem (\textbf{contribution~1}). Common practice in many modern RAG designs is to use embeddings derived from \textit{last-layer decoder block activations} of a decoder-based \textit{embedding LLM} (a language model specifically fine-tuned to provide high-quality embeddings). Examples of such embedding models are \texttt{SFR-Embedding-Mistral} and \texttt{E5-Mistral-7B}. {Our key idea} is to extend this design by incorporating activations from the \textit{multi-head attention (MHA) modules of decoder blocks} as embedding sources (see Figure~\ref{fig:embedding-model}). This enables the representation of multiple distinct aspects of the input text. Specifically, a Transformer consists of a stack of blocks (e.g., 96 in GPT-3~\citep{chu2024history}), each containing an MHA module with multiple \textit{heads} that are trained with separate parameter sets. Through a survey of literature on Transformer design and interpretability, we find empirical and theoretical support for the conjecture that \textit{different heads specialize in different aspects of the input}. This enables efficient \textit{multi-aspect embeddings} without increasing space or compute costs compared to standard RAG, and without requiring \textit{any} additional fine-tuning or architectural modifications to the base model.

\iftr
Considering multi-aspectuality comes with challenges. For example, it is unclear how to assess whether a RAG solution does indeed harness multiple aspects when fetching documents. For this, we develop a multi-aspect RAG pipeline that includes data preparation and query processing with both multi-aspect retrieval and ranking schemes (\textbf{contribution~2}). We also establish an evaluation methodology and provide multi-aspect datasets, complementing existing RAG benchmarks~\citep{chen2024benchmarking, xiong2024benchmarking, lyu2024crud, es2023ragas} (\textbf{contribution~3}). We ensure the relevance of our RAG datasets in real use cases by working directly with tech leaders (e.g., a generative AI division head) from 3 corporations, all of which actively use RAG in their own LLM infrastructures.
\fi
We illustrate the advantages of \nameAS over 18 traditional and modern RAG designs in various design criteria and in both time and space complexities (\textbf{contribution~4}).
In evaluation, \nameAS enhances the relevance of retrieved documents by up to 20\% over modern RAG baselines, offers comparable performance without degradation for single-aspect queries, and benefits the downstream LLM generation (\textbf{contribution~5}).
Due to its simplicity, \nameAS can be seamlessly integrated into any store while its benchmarking methodology can straightforwardly extend benchmarks such as RAGAs (\textbf{contribution~6}).
\ifnbld
\nameA's code is publicly available\footnote{\url{https://github.com/spcl/MRAG}}.
\fi


\section{\nameA: Design \& Implementation}
\label{sec:mrag}

\begin{figure*}[t]
  \centering
  \includegraphics[width=1.0\linewidth]{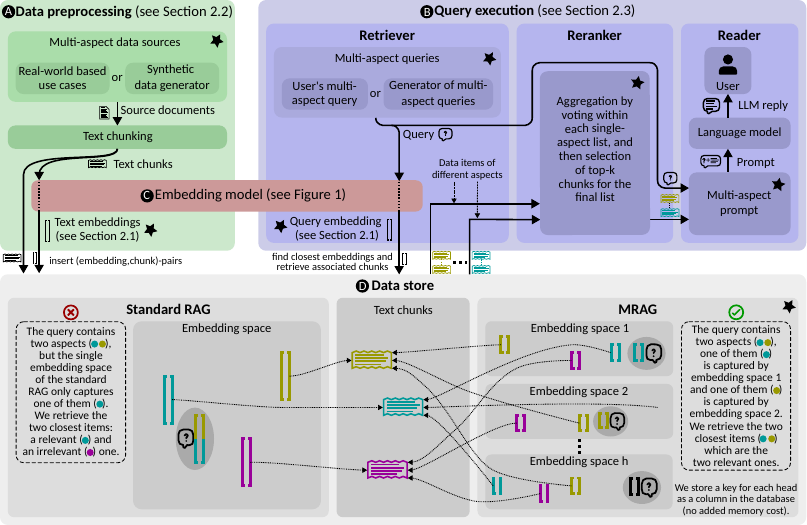}
  \caption{Overview of the \nameAS pipeline, consisting of two parts: data preparation~\iparta{} and query execution~\ipartb. The embedding model~\ipartc{} and the data store~\ipartd{} are used by both parts. The data store~\ipartd{} contains text embeddings~\iemb{} linking to text chunks~\ichunk{} reflecting three different aspects (\textcolor{c_cyan}{cyan}, \textcolor{c_magenta}{magenta}, \textcolor{c_yellow}{yellow}). Blocks marked by a star~\istar{} are a novelty of this work.}
  \vspaceSQ{-1.0em}
  \label{fig:overview}
\end{figure*}

A typical RAG scheme (see Figure~\ref{fig:overview}) consists of two main parts: \textbf{data preprocessing}~\iparta{} and \textbf{query execution}~\ipartb; both parts heavily use an \textbf{embedding model}~\ipartc{} and the \textbf{data store}~\ipartd. During preprocessing, each document in the database is encoded into one or more embeddings using the embedding model; these embeddings are stored with that document (usually as a key-value pair). During query execution, the embedding of the user-provided query is constructed using the same embedding model; then, the \textit{retriever} fetches candidate documents based on embedding similarity, the \textit{reranker} refines their ordering using more precise scoring, and the \textit{reader} uses a downstream generative model to synthesize the final answer from the top-ranked documents.

\subsection{Constructing Multi-Aspect Embeddings \iemb}\label{sec:multi-aspect-embedding}

An embedding is constructed for each data item in a pre-existing database (part~\iparta) and for the user query (part~\ipartb).
In standard modern RAG, given an input chunk of $n$ tokens constituting a document or chunk to be embedded, one obtains a corresponding embedding by applying the embedding model to the data item and extracting the activation of the final feed-forward layer for the last token $\mathbf{x}_n$, i.e., $\text{FFN}\left( \text{MHA}(\mathbf{x}_n) \right) \in \mathbb{R}^{d}$.
\iftr
Appendix~\ref{sec:appendix-maths} provides additional mathematical details.
\fi

In \nameA, instead of relying on this single embedding, the activations of all $h$ attention heads are obtained \textit{before} they are merged by the final projection layer. Specifically, for the last token $\mathbf{x}_n$, we extract the set of head-specific vectors $\mathcal{S} = \{\mathbf{e}_k\}_{k=1}^{h}$, where each $\mathbf{e}_k = \text{head}^k(\mathbf{x}_n) \in \mathbb{R}^{d / h}$ is referred to as a ``\textit{single-aspect embedding}'' and $\mathcal{S}$ is called a ``\textit{multi-aspect embedding}''. This provides $h$ semantically diverse embeddings per input, reflecting the different perspectives captured by each attention head. Since these vectors are extracted from the same internal activation, the space and time requirements remain unchanged from standard RAG. 
\iftr
However, the explicit separation of heads allows \nameAS to preserve fine-grained aspectual distinctions, facilitating multi-aspect representation.
\fi

\subsection{Data Preprocessing \iparta}

We populate a data store~\ipartd{} with multi-aspect embeddings~\iemb{} for the corresponding documents~\idoc{} or text chunks~\ichunk.
\iftr
\nameAS is orthogonal to the type of data being embedded, and while we primarily use chunking of documents in order to reflect modern RAG pipelines, one can also embed whole documents or even other types of data. 
\fi
Unlike in standard RAG, where a single embedding \iemb~points to a single text chunk \ichunk, in \nameA, each out of $h$ \textit{single-aspect} embeddings \iemb~points to the original text chunk \ichunk~(i.e., the data store \ipartd~contains $h$ embedding spaces). This crucial feature allows \nameAS to compare query \iquery~and text chunks \ichunk~in multiple embedding spaces that capture multiple aspects of the data.

\nameA\ performs one-time \textit{importance scoring} for each embedding space $i$, capturing its relevance to the dataset. Each score $s_i$ combines: (i) the average L2 norm $a_i$ of embeddings in space $i$, reflecting \textit{attention strength}, and (ii) the average cosine distance $b_i$ between sampled embeddings, approximating its \textit{semantic spread}. The final score is $s_i = a_i \cdot b_i$, encouraging both head relevance (through high $a_i$) and high representational diversity (through high $b_i$). Full details are provided in Algorithm~\ref{algo:indexing}, Appendix~\ref{sec:app:system-dets:scoring}.

\subsection{Query Execution}
\label{sec:query-exec}

During query execution, \nameAS first generates a {multi-aspect embedding $\mathcal{S}$} of the input query (cf.~Section~\ref{sec:multi-aspect-embedding}). These embeddings are computed in parallel during the same inference pass, so this multi-vector representation introduces {no additional latency}.

\textbf{Retriever.} Given $\mathcal{S}$, the retriever retrieves in {parallel} the top-$c$ nearest chunks within each embedding space, and then {aggregates} the resulting candidates across all $h$ spaces.

\textbf{Reranker.} Once $hc$ candidate chunks are retrieved across all heads, the {ranker} stage consolidates them into a single list of top-$k$ results using a simple but effective {voting strategy}. For each candidate chunk at position $p$ in the ranked list for space $i$, we assign a score of $s_i \cdot 2^{-p}$, where $s_i$ is the precomputed importance score. This exponentially discounts lower-ranked candidates and balances influence across heads. The final top-$k$ list is obtained by globally sorting all candidates by these scores. The voting procedure is described in Algorithm~\ref{algo:voting} in Appendix~\ref{sec:app:system-dets:ranking}.

Our voting-based ranker is intentionally lightweight, adding negligible overhead beyond sorting (Algorithm~\ref{algo:voting}). At the same time, \nameA\ is \emph{reranker-agnostic}: it outputs a standard top-$k$ candidate set that can (if needed) be seamlessly passed to any downstream reranker (e.g., a cross-encoder) without changing the retriever or index. We do not include interaction-based rerankers in our default pipeline since they require \textit{pairwise} query--document scoring, incurring $\Theta(k)$ additional model evaluations (each with typically full self-attention cost)~\citep{nogueira2020passage, sbert_retrieve_rerank}.

\textbf{Reader.} The top-$k$ retrieved results are inserted into the LLM context using a multi-aspect prompt template (prompts are fully specified in Appendix~\ref{sec:app:prompts}). Each result is placed in a separate section of the prompt. Stored metadata can be included alongside each chunk to provide additional context, such as the aspect or the chunk category. 

\subsection{Multi-Aspectuality Without Additional Training}
\label{sec:multiaspects-no-training}

We specifically use \textit{no additional fine-tuning}, to avoid training overhead and make deployment easy. This is motivated by growing evidence that attention heads in Transformer models naturally converge during training to focus on distinct aspects of the input. For example, \citet{wang2024differentiation} show that heads diverge into clusters specialized for different input patterns, while \citet{olsson2022incontext} and others~\citep{mcdougall2024copy, gould2024successor} discover that heads focus on repeated sequences or named entities. Similar trends are observed in BERT~\citep{clark2019bert, kovaleva2019revealing, htut2019attention}. We provide more details in a brief literature survey (Appendix~\ref{sec:app:heads-survey}).
Given these findings, we assume that even without additional fine-tuning, the embeddings output by MHA already encodes multi-aspectuality suitable for downstream retrieval. Our own brief attention pattern analysis (Appendix~\ref{sec:app:heads-analysis}) confirms shifting token focus across heads.

\iftr


\begin{table*}[ht]
\setlength{\tabcolsep}{1.5pt}
\ifsq\renewcommand{\arraystretch}{0.6}\fi
\centering
\footnotesize
\scriptsize
\caption{\textbf{Time and storage complexity of different RAG schemes; only \nameAS matches the asymptotic complexity of vanilla RAG}.
\textbf{Work} is the total number of all operations, \textbf{Depth} is runtime complexity assuming unlimited parallel processing units; both are established measures of analyzing parallel algorithms.
\textbf{Retrieval} describes steps during inference, including any postprocessing before returning documents or summaries.
\textbf{Preprocessing} includes all steps before inference.
$n$: number of data items in a database (i.e., document chunks), 
$d$: embedding dimensionality, 
%
%
$k$: number of retrieved top chunks per single user query, 
$l_q / l_d$: average token length of the query/document, 
$W_m$/$D_m$: work/depth to run a Transformer-based model,
$W_e$/$D_e$: work/depth to embed a graph,
$W_i$/$D_i$: work/depth to index a graph,
$s$: polynomial function that models the complexity of various additional scheme-specific design decisions, heuristics, etc., which cannot be straightforwardly modeled with closed-form expressions (e.g., count of graph communities, toy-graph size, self-note length, BM25 matching cost, keyword matching cost); it scales at most linearly with $n$ and is typically considerably larger than $O(1)$.}
\label{tab:complexity}
\resizebox{\textwidth}{!}{%
\begin{tabular}{@{}lllllll@{}}
\toprule
\multirow{2}{*}{\textbf{Scheme}} & \multicolumn{2}{l}{\textbf{Retrieval}} & & \multicolumn{2}{l}{\textbf{Preprocessing}} & \multirow{2}{*}{\,\textbf{Storage}} \\
 \cmidrule{2-3} \cmidrule{5-6} 
 & \textbf{Work} & \textbf{Depth} & & \textbf{Work} & \textbf{Depth} &  \\
\midrule \textbf{[S] Dense single-vector} \\ \midrule
\quad \textbf{Vanilla RAG} & $O(W_m + n d)$ & $O(D_m + \log d)$ & & $O(W_m n)$ & $O(D_m)$ & $O(n d)$ \\ 
\quad \citet{lewis2020retrieval} & $O(2W_m + n d)$ & $O(2D_m + \log d)$ & & $O(W_m n)$ & $O(D_m)$ & $O(n d)$ \\
\midrule \textbf{[M] Dense multi-vector} \\ \midrule
\quad \textbf{\nameAS [This Work]} & $O(W_m + n d)$ & $O(D_m + \log d)$ & & $O(W_m n)$ & $O(D_m)$ & $O(n d)$ \\
\quad Poly-encoders~\citep{humeau2019poly} & $O(W_m + nd + s(n)d)$ & $O(D_m + \log d)$ &   & $O(W_m n)$ & $O(D_m)$ & $O(n d)$ \\
\quad ColBERT~\citep{khattab2020colbert} & $O(W_m + n d l_{q} l_{d})$ & $O(D_m + \log dl_d)$ & & $O(W_m n)$ & $O(D_m)$ & $O(n dl_d)$ \\
\midrule \textbf{[Q] Query expansion}\\\midrule
\quad RQ-RAG~\citep{chan2024rq} & $O(s(n)(W_m+nd))$ & $O(s(n)(D_m + \log d))$ &  & $O(W_m n)$ & $O(D_m)$ & $O(nd)$ \\
\quad Fusion RAG~\citep{rackauckas2024rag} & $O(s(n)W_m+knd)$ & $O(2D_m+\log d)$ & & $O(W_m n)$ & $O(D_m)$ & $O(nd)$ \\
\midrule\textbf{[I] Iterative / agentic} \\\midrule
\quad EMDR~\citep{singh2021end} & $O((k+1)W_m + n d)$ & $O(2D_m + \log d)$ & & $O(W_m n)$ & $O(D_m)$ & $O(n d)$ \\
\quad Self-RAG~\citep{asai2023selfrag} & $O((2k+1)W_m + n d + k)$ & $O(2D_m + \log d)$ & & $O(W_m n)$ & $O(D_m)$ & $O(n d)$ \\
\quad Chain-of-Note~\citep{yu2023chain} & $O((ks(n)+1)W_m + n d)$ & $O((ks(n)+1)D_m + \log d)$ &  & $O(W_m n)$ & $O(D_m)$ & $O(n d)$ \\
\quad ThinkNote~\citep{xu2024active} & $O(3W_m + nd)$ & $O(2D_m+\log n)$ & & $O(W_m n)$ & $O(D_m)$ & $O(n d)$ \\
\midrule\textbf{[H] Structure-/hierarchy-augmented} \\\midrule
\quad RAPTOR~\citep{sarthi2024raptor} & $O(W_m + n d + s(n) d)$ & $O(D_m + \log d)$ & & $O(W_m n + W_m s(n))$ & $O(D_m)$ & $O(n d + s(n) d)$ \\
\quad RAGraph~\citep{jiang2024ragraph} & $O(W_m + n d + s(n) d)$ & $O(D_m + \log d)$ & & $O(W_m + W_e)$ & $O(D_m+D_e)$ & $O(n d + s(n) d)$ \\
\quad HiQA~\citep{chen2024hiqa} & $O(W_m+nd+s(n)d)$ & $O(D_m+\log d)$ &  & $O(W_m n + s(n)d)$ & $O(D_m)$ & $O(n d+s(n)d)$ \\
\quad GraphRAG~\citep{edge2024from} & $O(s(n)(W_m+d))$ & $O(D_m+s(n)d)$ &  & $O(W_m n + W_ms(n) + W_e)$ & $O(D_m+D_e)$ & $O(n d+s(n)d)$ \\
\quad SuperRAG~\citep{yang2025superrag} & $O(W_m + n d + W_i)$ & $O(D_m + \log d +D_i)$ &  & $O(W_m n + s(n) + W_e)$ & $O(D_m+D_e)$ & $O(n d+s(n))$ \\
\quad HiRAG~\citep{huang2025retrieval} & $O(W_m + n d + W_i)$ & $O(D_m + \log d +D_i)$ &  & $O(W_m n + s(n) + W_e)$ & $O(D_m+D_e)$ & $O(n d+s(n))$ \\
\midrule\textbf{[C] Chunking / context organization} \\\midrule
\quad Meta-chunking~\citep{zhao2024meta} & $O(W_m + (n+s(n))d)$ & $O(D_m + \log d)$ &  & $O(l_dW_m+s(n)W_m)$ & $O((l_d+1)D_m)$ & $O(nd+s(n)d)$ \\
\quad MoC~\citep{zhao2025moc} & $O(W_m + (n+s(n))d)$ & $O(D_m + \log d)$ &  & $O(W_mn+s(n))$ & $O(D_m)$ & $O(nd+s(n)d)$ \\
\midrule\textbf{[P] Parametric}\\\midrule
\quad Parametric RAG~\citep{su2025parametric} &  $O(W_m + n d + s(n))$ & $O(D_m + \log d +s(n))$ & & $O(W_m n s(n))$ & $O(D_ms(n))$ & $O(ns(n))$ \\
\bottomrule
\end{tabular}
}
%
\vspaceSQ{-1em}
\end{table*}

\fi

\subsection{Parallelization and Lightweight System Design}
\label{sec:parallelization}

MRAG keeps the total storage and compute the same as standard RAG ($O(nd)$; we show a more detailed complexity analysis in the following section). Importantly, \nameA\ does not increase embedding latency: all $h$ head vectors are extracted from a \emph{single} forward pass of the embedding model. Retrieval then performs $h$ independent ANN lookups (one per head space) and merges $hc$ candidates; this is embarrassingly parallel and can be executed as batched / concurrent ANN queries. We use off-the-shelf ANN indexes with parallel subspace search, and avoid dynamic or variable-length vectors that complicate indexing and caching. 

Modern vector databases already support the parallel multi-vector search required by our approach. Milvus can execute parallel ANN queries across shards with low latency \citep{wang2025towards,clavie2024reducing,wang2021milvus}. Pinecone provides cascading retrieval with fixed-size multi-vector encodings to maintain predictable costs. ESPN tackles multi-vector retrieval at SSD and GPU levels by implementing prefetching pipelines and storage bypass mechanisms, achieving near-memory query latency, even when dealing with large indices \citep{shrestha2024espn}. These technologies illustrate that multi-vector retrieval, especially when using fixed-size vectors as MRAG does, can be implemented at scale with negligible practical overhead. 

\section{Compute \& Storage Complexity Analysis}
\label{sec:complexity}

We analyze the runtime and space complexity of \nameAS and 18 RAG baselines in Table~\ref{tab:complexity}. Derivation details are in Appendix~\ref{sec:app:complexity}.
Overall, \nameAS achieves competitive results in all aspects. At inference time, it extracts $h$ attention-head embeddings in parallel from a single forward pass, resulting in the same latency as standard RAG. Preprocessing is lightweight, requiring only one pass per document and simple statistics for head scoring, avoiding the cost of training or complex structure construction. Storage overhead is minimal, as $h$ single-aspect embeddings per data item have the same dimensionality as a standard embedding. In contrast, prior schemes like Poly-encoder~\citep{humeau2019poly} and ColBERT~\citep{khattab2020colbert} incur significant cost due to using many token-level embeddings or time-consuming training rounds.

\section{Benchmarking Multi-Aspectuality}
\label{sec:multi-aspect-data-and-metrics}
\enlargeSQ

For evaluation, we need (1) datasets of multi-aspect documents, (2) queries to the LLM that require retrieving multi-aspect documents, and (3) metrics that assess how well a RAG scheme retrieves such multi-aspect data. We summarize below these three elements; details are in Appendix~\ref{sec:app:multi-aspect-data-and-metrics}.

We construct three \textbf{multi-aspect datasets}: (1) a Wikipedia dataset with documents sampled from clearly distinct categories (e.g., countries, shipwrecks, board games); (2) a real-world-based legal document dataset with aspects spanning different legal areas (energy law, family law, criminal law, etc.) and document language style (aggressive, mild, neutral, etc.); and (3) a real-world-based dataset of industry accident reports, categorized by cause. For each dataset, we generate \textbf{multi-aspect queries} combining $n$ distinct categories into single queries (with $n \in \{1, ..., 25\}$). For example, a query with 10 aspects must contain a question about 10 distinct documents from 10 different categories.

\begin{figure*}[t]
    \centering
        \includegraphics[width=1.0\linewidth]{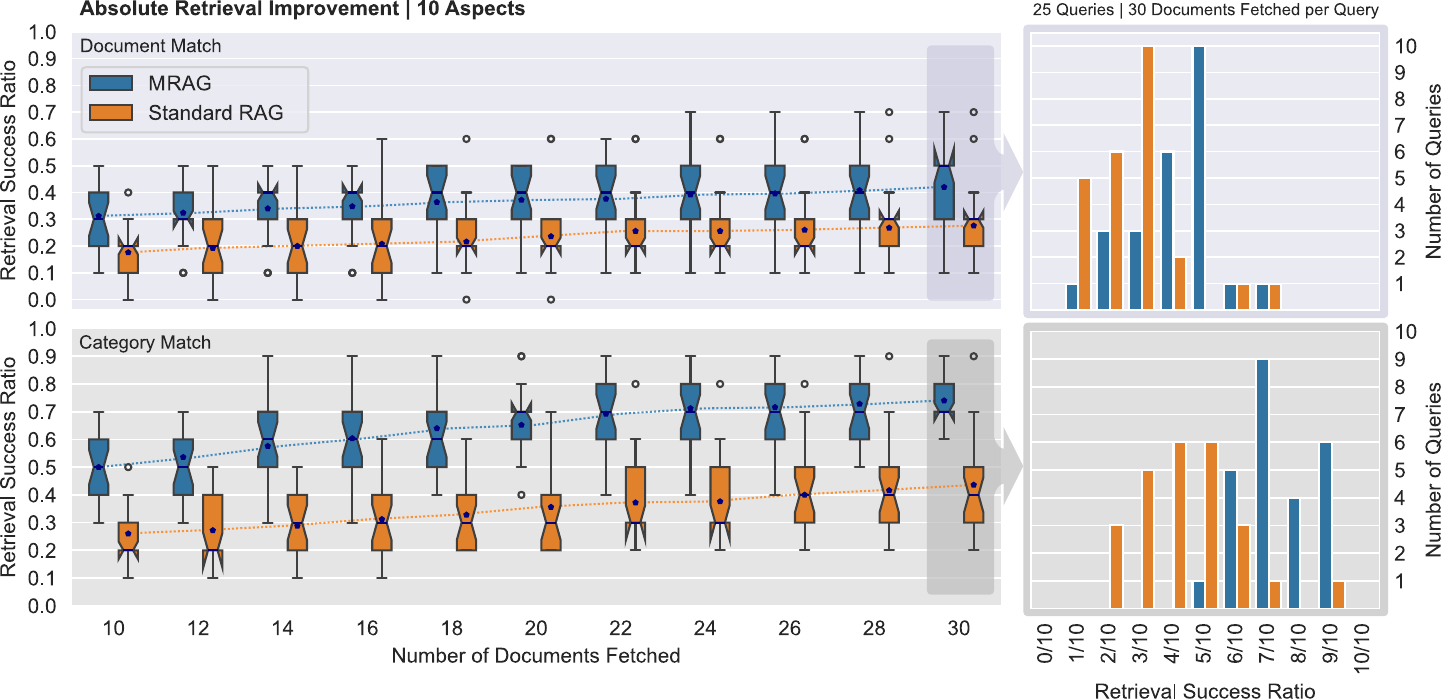}
    \caption{\textbf{Retrieval success ratio (25 queries) between \nameAS and Standard RAG}; each query uses 10 aspects. The top part presents \textbf{exact document matches}, the bottom part presents \textbf{category only matches} (we explain the metrics in Sec.~\ref{sec:multi-aspect-data-and-metrics}). A histogram is presented for a specific sample to showcase the detailed distribution among the 25 queries (there are 30 documents fetched for \textit{each} query).}
    \vspaceSQ{-1.0em}
    \label{fig:plot_qt10_absolute}
\end{figure*}

To evaluate retrieval performance, we introduce three \textbf{metrics for assessing multi-aspectuality}. Let $Q$ be a query, $S$ a reranker scheme, and $Q_{rel}$ the ideal set of $n$ relevant documents. The \textit{Retrieval Success Ratio} is defined as $\Xi(Q, n) = \frac{|S(Q, n) \cap Q_{rel}|}{|Q_{rel}|}$ (the fraction of exactly matched documents). The \textit{Category Retrieval Success Ratio}~$\Xi_c$ extends this by also accepting matches from the same categories as those in $Q_{rel}$ even if the exact document has not been matched. Finally, we define a tunable \textit{Weighted Success Ratio}~$\Xi_w=\frac{w\cdot\Xi+\Xi_c}{w+1}$, allowing the user to adjust the trade-off between exact and category-level matches via $w$.

\textbf{Why Category-Level Matches?}
We motivate category-level retrieval using both previous works~\citep{venktesh2025sunar, liu2021dense, chen2024class, xiaoorag} as well as our hands-on experience in legal and industrial accident analysis. Even when exact matches are missing, retrieving documents from the same semantic category can enhance generation, as supported by the classic \textit{clustering hypothesis} in information retrieval~\citep{venktesh2025sunar, liu2021dense}: documents in the same ``neighborhood'' are likely to be relevant to the same query. Recent studies further validate this across open-domain QA and ontology-guided RAG~\citep{chen2024class, xiaoorag}.

\section{Evaluation}
\label{sec:eval}

\enlargeSQ

We now illustrate the advantages of \nameAS over the state of the art. Further details on evaluation setup and additional results are in -- respectively -- Appendix~\ref{sec:app:setup} and~\ref{sec:app:eval}.


\textbf{\underline{Comparison Baselines.}}
We consider three main baselines: \textbf{Standard RAG}, \textbf{Split RAG}, and \textbf{Fusion RAG}~\citep{rackauckas2024rag}. The first represents a modern RAG pipeline in which each document uses the activations of the last decoder layer as its embedding. The second is a blend between Standard RAG and \nameA; it splits the activation of the last decoder layer in the same way as \nameAS and applies a voting strategy. The purpose of {Split RAG} is to show that \textit{\nameA's benefits come from using the multi-head output as embedding and \textbf{not} merely using multiple embedding spaces}.
%
%
\textbf{Fusion RAG}~\citep{rackauckas2024rag} is an optional mechanism that we harness to \textit{further enhance the benefits of \nameAS at the tradeoff of additional tokens} (detailed in Section~\ref{sec:fusion-rag}). 
%

\textbf{\underline{Models. }}
While \nameAS can extract multi-aspect embeddings from \textit{any} block, we found that the last MHA works best.
\iftr
\nameAS can use any embedding model with MHA, including models such as RetroMAE~\citep{xiao2022retromae} and the classic BGE-embeddings~\citep{xiao2022retromae, chen2024bge}. We consider two embedding models from the MTEB leaderboard~\citep{mteb2024, muennighoff2023mteb}, \texttt{SFR-Embedding-Model} (SFR)~\citep{SFRAIResearch2024} and \texttt{e5-mistral-7b-instruct} (e5)~\citep{wang2022text}, both based on the Mistral 7B architecture with 32 decoder blocks and attention heads.
\fi

\iftr
We use \textbf{queries} and \textbf{metrics} introduced in Section~\ref{sec:multi-aspect-data-and-metrics}. We use the weighted retrieval success ratio with 2:1 weighting, which considers category matches as relevant but prioritizes exact document matches.
\fi

\iftr
\textbf{Samples \& Summaries}
Each data point in our plots corresponds to 25 queries. We present the data using standard boxplots to showcase the distribution. Our primary focus is on the average retrieval performance among those 25 queries. 
\fi

\enlargeSQ

\begin{figure*}[t]
    \centering
        \includegraphics[width=1.0\linewidth]{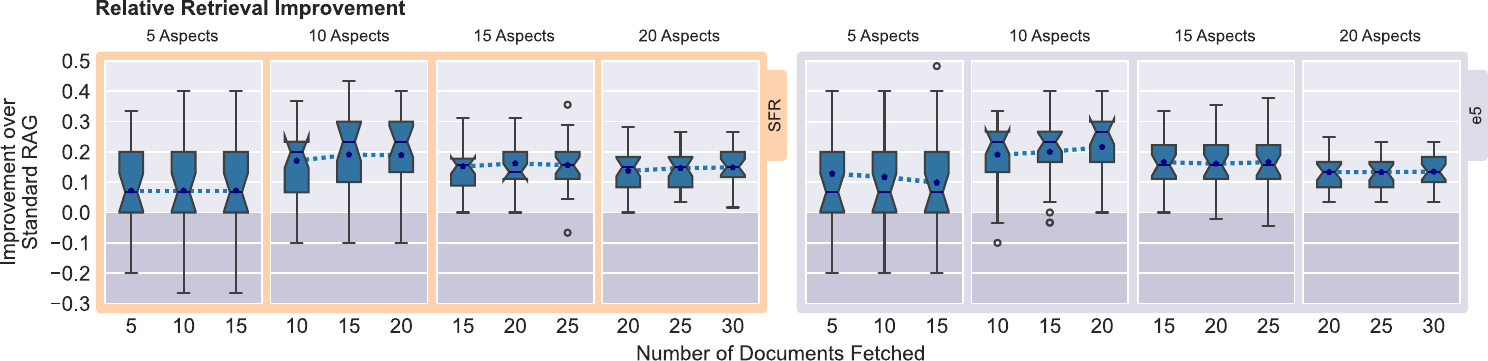}
        \vspace{-1.50em}
    \caption{\textbf{Relative retrieval improvement of \nameAS over Standard RAG} across queries with different numbers of aspects and different embedding models (left side: SFR, right side: e5).}
    \vspaceSQ{-1.0em}
    \label{fig:plot_relative}
\end{figure*}

\iftr
\begin{figure*}[t]
    \centering
        \includegraphics[width=0.8\linewidth]{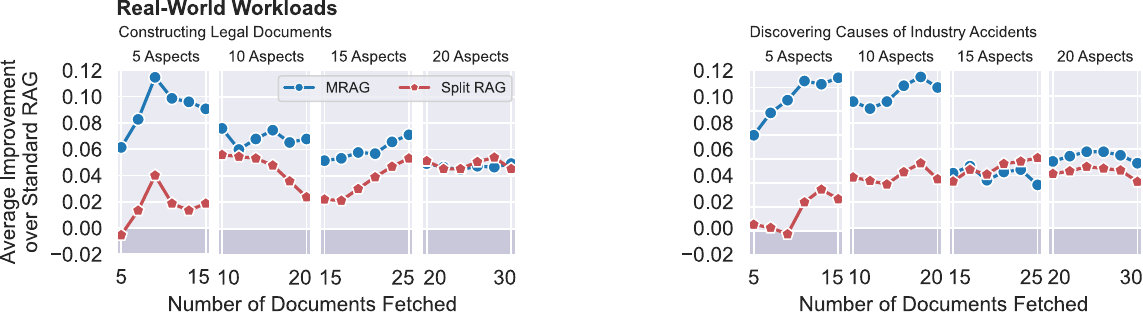}
    \caption{\textbf{Average improvement of the retrieval success ratio of \nameAS and Split RAG over Standard RAG} for two real-world workloads \textit{constructing legal documents} (left) and \textit{discovering causes of industry accidents} (right).}
    \label{fig:plot_real}
\end{figure*}
\fi

\subsection{Superior Performance for Multi-Aspect Queries}
\label{sec:eval_analysis}

We begin with the {Wikipedia multi-aspect dataset} (cf.~Section~\ref{sec:multi-aspect-data-and-metrics}). In each query, we mention the documents to be fetched in the text and then assess the success ratio of different RAG strategies in finding these documents and their categories (a full example of such a query is in Figure~\ref{fig:query_detail} in Appendix~\ref{sec:app:multi-aspect-data-and-metrics}).
We show first the absolute retrieval performance of \nameAS over Standard RAG in Figure~\ref{fig:plot_qt10_absolute}. We fix the number of aspects present in the queries to 10, and vary the total number of retrieved documents from 10 to 30. \nameAS consistently outperforms Standard RAG ($>10\%$ increase in the retrieval success ratio on average for exact document matches). Moreover, the retrieval performance increase is even more significant on category matches ($>25\%$ increase in the retrieval success ratio on average). The performance increase is further detailed in the histograms on the right side. Here, for a specific number of documents fetched, \nameA's histogram indicates a better distribution of retrieval success ratios (across all 25 queries).
This gain stems from \nameA's ability to decompose query semantics across multiple attention heads, increasing the likelihood of matching each aspect with a semantically aligned document.

%
Next, Figure~\ref{fig:plot_relative} shows the relative weighted performance improvement of \nameAS with respect to Standard RAG as we vary the number of aspects present in the queries. We show data for two different embedding models (SFR and e5). \nameAS consistently outperforms the Standard RAG by 10-20\% on average, not only across the number of documents fetched, but also across the increasing counts of aspects present in the replies, and does so for both embedding models.
%
This robustness suggests that \nameAS scales better than Standard RAG with query complexity, as its multi-head representation distributes semantic load more evenly than the single-vector baseline.

To further illustrate advantages of \nameA, we also consider two \textbf{real-world use cases} from in-house industry data analytics projects, namely, the synthesis of legal documents and the analysis of causes of chemical plant accidents. The results are in Figure~\ref{fig:plot_real}.
In the former (left side), the task is to create a document based on user requirements that may be related to different \textit{aspects}, for example to the law being considered (e.g., the British or the US one), the subject (e.g., energetic or civil), the style of the document (e.g., aggressive or mild), and others. This task is executed with RAG that can fetch documents from a database.
In the latter (right side), the task is to discover a cause of an accident. In this setting, the objective is to retrieve documents from a database that should be used in the LLM context to facilitate discovering the cause of the accident. The causes are grouped in categories such as utility impact due to severe weather, lack of preparedness and planning, incorrect installation of equipment, lack of maintenance, among others.
Similarly to the previous analyses, we measure the retrieval success ratio over corresponding databases. \nameAS offers advantages over other schemes.

\iftr
\begin{figure*}[t]
  \centering
  \includegraphics[width=1.0\linewidth]{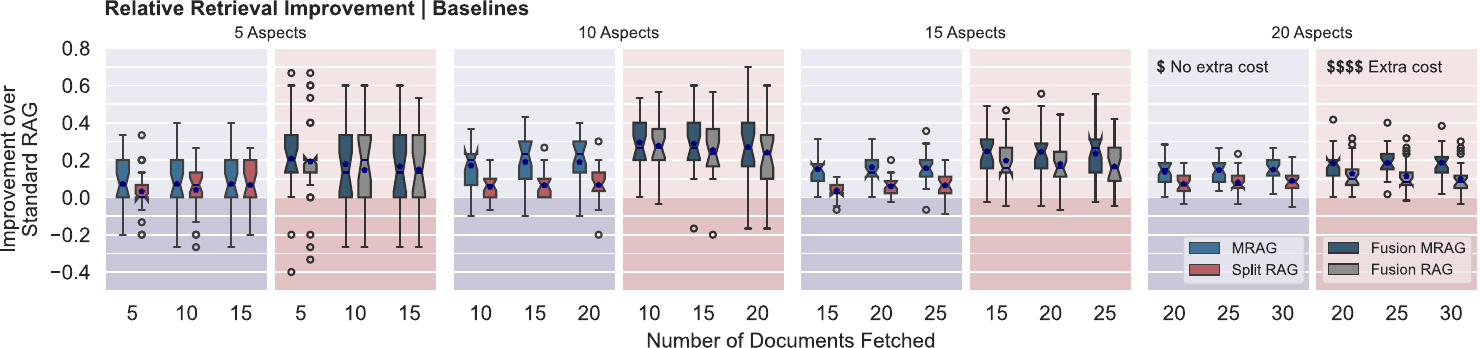}
  \caption{\textbf{Relative retrieval improvements of \nameAS over Standard RAG} for the SFR embedding model compared with \textbf{Split RAG} (blue plots), and the \textbf{relative retrieval improvements of Fusion \nameAS over Standard RAG} compared with \textbf{Fusion RAG} (red plots).}
  \label{fig:plot_relative_comparison}
\end{figure*}
\fi

\enlargeSQ

We also \textbf{delve deeper into the underlying factors for \nameA's performance gains}.
For this, we compare \nameAS to the Split RAG baseline in Figure~\ref{fig:plot_relative_comparison}. 
%
%
The blue plots show the relative weighted performance of \nameAS and Split RAG over Standard RAG. \nameAS performs better than Split RAG, illustrating that its \textit{high accuracy is due to the actual multi-head part}, and not merely just partitioning the vector and using multiple embedding spaces. 
%


\definecolor{my_gray}{HTML}{7e7e7e}

\subsection{High Performance Also for Single-Aspect Queries}
\label{sec:single_aspect}

We additionally show in Table~\ref{tab:query_type_1} that \nameAS performs on par with Standard RAG on queries where only a single aspect is expected. This confirms that \nameA's design generalizes well: in low-aspect settings, the aggregation over heads still captures dominant semantics, effectively collapsing into a strong single-vector representation 
with only a negligible accuracy drop for single-aspect tasks.

\begin{table}[t]
\vspaceSQ{-0.5em}
\centering
\setlength{\tabcolsep}{1.5pt}
\ifsq\renewcommand{\arraystretch}{0.7}\fi
%
\footnotesize
\scriptsize
\ssmall
\sf
\caption{\textbf{Retrieval success ratio} (the exact document match) for 25 \textbf{single-aspect} queries on different datasets (Multi-Aspect Dataset, Legal Dataset, Accidents Dataset), using different embedding models (SFR, e5). For every query, a specific single document (single-aspect) is expected to be among the fetched documents for the retrieval to be classified as successful.}
\vspaceSQ{-0.5em}
\label{tab:query_type_1}
\begin{tabular}{cccccccccccc}
\toprule
\multirow{3}{*}{\makecell[l]{\textbf{Documents} \\ \textbf{Fetched}}} & \multicolumn{5}{c}{{\textbf{Wikipedia Dataset}}} & & \multicolumn{2}{c}{{\textbf{Legal Dataset}}} & & \multicolumn{2}{c}{{\textbf{Accidents Dataset}}} \\

& \multicolumn{2}{c}{{\textbf{SFR}}} & & \multicolumn{2}{c}{{\textbf{e5}}} & & \multicolumn{2}{c}{{\textbf{SFR}}} & & \multicolumn{2}{c}{{\textbf{SFR}}} \\
\cmidrule(lr){2-12} 
& \textbf{\nameA} & \textbf{RAG} & | & \textbf{\nameA} & \textbf{RAG} & | & \textbf{\nameA} & \textbf{RAG} & | & \textbf{\nameA} & \textbf{RAG} \\
\midrule
1 & {\color{my_gray}24/25} & 25/25 & & {\color{my_gray}24/25} & 25/25 & & {\color{my_gray}24/25} & {\color{my_gray}24/25} & & 25/25 & 25/25 \\
2 & 25/25 & 25/25 & & 25/25 & 25/25 & & 25/25 & 25/25 & & 25/25 & 25/25 \\
3 & 25/25 & 25/25 & & 25/25 & 25/25 & & 25/25 & 25/25 & & 25/25 & 25/25 \\
\bottomrule
\end{tabular}
\end{table}


\subsection{Seamless Enhancement of Existing RAG Schemes}
\label{sec:fusion-rag}

\nameA's simplicity ensures that it can be seamlessly integrated with other RAG approaches. As an example, we combine \nameAS with \textit{Fusion RAG}, which uses an LLM (additional token cost) for even more accurate retrieval.
Fusion RAG uses an LLM to create a fixed number of questions about the RAG query. Each question is separately applied through an embedding model using Standard RAG. This relies on multiple LLM calls and heuristic reranking, inflating latency and cost.
We apply \nameAS to each of these questions and denote the combined scheme as \textit{Fusion \nameA}. 
Red plots of Figure~\ref{fig:plot_relative_comparison} show that Fusion \nameAS consistently outperforms pure Fusion RAG, indicating that these optimizations can be combined together. However, both Fusion strategies introduce a greater variance than \nameAS and additional costs in terms of compute, latency, and tokens.

%
%

\subsection{Advantages for Downstream LLM Generation}

We also show that \nameAS enhances the downstream LLM generation with its improved multi-aspect retrieval. For this, we use a sampled subset of multi-aspect Wikipedia queries, for which we applied both Standard RAG and \nameAS to retrieve supporting documents. These documents are then integrated into the prompt templates and are passed to the LLM to answer the original query with the aid of the retrieved data. To quantify the effect from RAG, we count facts in the LLM output (e.g., years or named entities such as locations). On average, \nameAS generations contain 15.4 pieces of factual information per query, compared to 11.2 for Standard RAG. The average improvement further confirms \nameA's advantages. By explicitly leveraging multi-aspect embeddings from MHA, one increases the likelihood that the LLM generation includes diverse and complementary facts, especially for complex queries spanning multiple domains (e.g., combining historical events with technological timelines). Overall, \nameAS helps the LLM in delivering richer, more comprehensive responses.

\subsection{No Latency and Storage Overheads}

\nameAS introduces no latency overhead at query embedding time, as all head-level embeddings are extracted in parallel from a \textit{single} standard forward pass. Retrieval across embedding subspaces is also fully parallelizable with modern vector databases~\citep{pan2024vector, han2023comprehensive}, and our use of a modest number of heads (e.g., 16–32 for \texttt{SFR-Embedding-Model}) ensures this parallelism is within easy reach. As the total dimensionality of embeddings remains unchanged (e.g., 1024 split across 32 heads), \nameAS also needs no additional storage compared to standard RAG.

\enlargeSQ

\subsection{Confirmation of Specialization of Heads}

\begin{figure}[t]
    \centering
        \includegraphics[width=0.95\columnwidth]{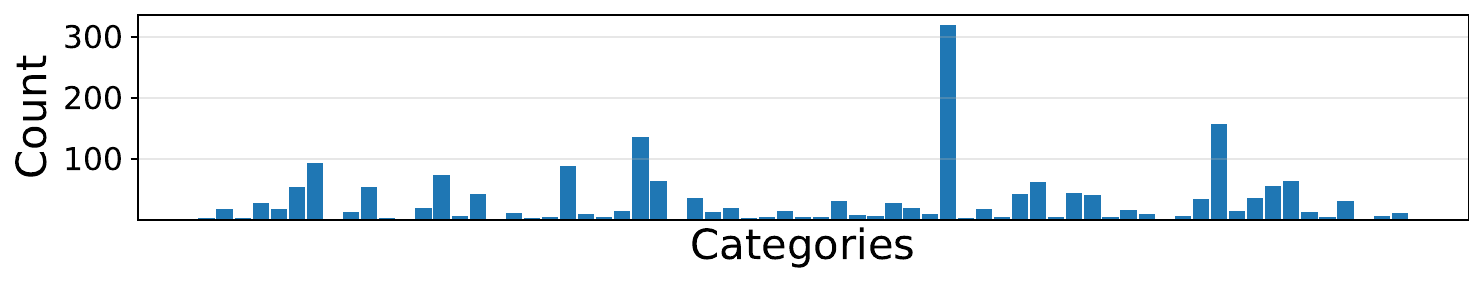}
        \vspaceSQ{-0.5em}
    \caption{{Higher-than-average retrieval ratio of documents from the \textit{mathematical transforms} category by head~7 (highest bar)}.}
    \vspaceSQ{-0.5em}
    \label{fig:head-spec}
\end{figure}

\ifconf
\else
We observe that different heads do indeed retrieve disproportionately more documents for certain aspects (more than 2$\times$ higher than average for some head--category pairs). For example, Figure~\ref{fig:head-spec} shows that retrievals from the \textit{mathematical transforms} category dominate head~7 for the Wikipedia multi-aspect dataset. Further analyses are in Appendix~\ref{sec:app:heads-analysis}).
\fi

\iftr
\subsection{Further Analyses \& Ablation Studies}
\label{sec:limitations}

Additional analyses are in Appendix~\ref{sec:app:eval}. They include analyzing the impact of using embeddings from \textbf{different decoder blocks} (rather than the last one), \textbf{scalability of preprocessing}, and \textbf{additional voting strategies for reranking}. These analyses all confirm the previous findings of \nameAS broadly outperforming other baselines.
\fi

\begin{table*}[t]
\centering
\ifconf
\else
\scriptsize
\fi
\setlength{\tabcolsep}{3pt}
\ifsq\renewcommand{\arraystretch}{0.6}\fi
\caption{\textbf{\ul{Comparison of the advantages of different RAG schemes}} (sorted chronologically).
\textbf{Type} follows the same categorization as in Table~\ref{tab:complexity}.
\textbf{No additional training}: does a given scheme require additional training beyond standard pre-training and fine-tuning?
\textbf{Works with any MHA LLM}: does a given scheme work seamlessly with any multi-head attention (MHA) LLM?
\textbf{Extensibility to other modalities}: could a given scheme be relatively easily used beyond LLMs, e.g., with Graph Foundation Models (GFMs), Vision Models, and others?
\textbf{No overhead at inference}: does a given scheme enable zero additional overhead at inference?
\textbf{Scalable preprocessing}, \textbf{low storage overhead}: does a given scheme enable scalable preprocessing and little storage overhead, respectively? (details in Section~\ref{sec:complexity} and Table~\ref{tab:complexity}).
\textbf{Multi-Aspectuality}: does a given scheme enable extracting multiple aspects of indexed data?
``\faY'': full support, ``\faH'': partial support, ``\faN'': no support, ``\faU'': unknown.
}
\vspaceSQ{-0.5em}
\label{tab:qualitative}
\begin{tabular}{@{}lllllllllll@{}}
\toprule
\texttt{\textbf{[Type]}}\,\textbf{Scheme} & \makecell[c]{\textbf{No additional} \\ \textbf{training}} & \makecell[c]{\textbf{Works with any} \\ \textbf{MHA LLM}} & \makecell[c]{\textbf{Extensibility to} \\ \textbf{other modalities}} & \makecell[c]{\textbf{No overhead} \\ \textbf{at inference}} & \makecell[c]{\textbf{Scalable} \\ \textbf{preprocessing}} & \makecell[c]{\textbf{Low storage} \\ \textbf{overhead}} & \makecell[c]{\textbf{Multi-} \\ \textbf{Aspectuality}} \\
\midrule
\texttt{\textbf{[S]}} \citet{lewis2020retrieval} & \faN & \faY & \faN & \faN & \faY & \faH &  \faN \\
\texttt{\textbf{[M]}} Poly-encoders~\citep{humeau2019poly} & \faN & \faN\ (BERT based) & \faN & \faN & \faY & \faH & \faN \\
\texttt{\textbf{[M]}} ColBERT~\citep{khattab2020colbert} & \faN & \faN\ (BERT based) & \faN & \faH & \faY & \faH & \faH \\
\texttt{\textbf{[I]}} EMDR~\citep{singh2021end} & \faN & \faY & \faN & \faY & \faH & \faH &  \faN \\
\texttt{\textbf{[I]}} Self-RAG~\citep{asai2023selfrag} & \faN & \faY & \faN & \faN & \faN & \faN &  \faN \\ 
\texttt{\textbf{[I]}} Chain-of-Note~\citep{yu2023chain} & \faN & \faH & \faN & \faN & \faY & \faH & \faN \\
\texttt{\textbf{[I]}} ThinkNote~\citep{xu2024active} & \faY & \faY & \faN & \faN & \faH & \faH & \faN \\
\texttt{\textbf{[H]}} RAPTOR~\citep{sarthi2024raptor} & \faY & \faY & \faN & \faN & \faY & \faH & \faN \\
\texttt{\textbf{[H]}} RAGraph~\citep{jiang2024ragraph} & \faY & \faN\ (only GFMs) & \faN\ (only GFMs) & \faH & \faH & \faN &  \faH \\
\texttt{\textbf{[H]}} HiQA~\citep{chen2024hiqa} & \faY & \faY & \faN & \faN & \faH & \faN & \faH \\
\texttt{\textbf{[H]}} GraphRAG~\citep{edge2024from} & \faY & \faY & \faU & \faN & \faY & \faY & \faH \\
\texttt{\textbf{[H]}} SuperRAG~\citep{yang2025superrag} & \faY & \faH & \faY & \faN & \faH & \faH & \faN \\
\texttt{\textbf{[H]}} HiRAG~\citep{huang2025retrieval} & \faY & \faY & \faU & \faN & \faY & \faH & \faN \\
\texttt{\textbf{[Q]}} RQ-RAG~\citep{chan2024rq} & \faN & \faH & \faN & \faN & \faY & \faY & \faN \\
\texttt{\textbf{[Q]}} Fusion RAG~\citep{rackauckas2024rag} & \faY & \faH & \faN & \faN & \faY & \faY & \faN \\
\texttt{\textbf{[C]}} Meta-chunking~\citep{zhao2024meta} & \faY & \faH & \faN & \faY & \faY & \faH & \faN \\
\texttt{\textbf{[C]}} MoC~\citep{zhao2025moc} & \faN & \faH & \faN & \faN & \faY & \faY & \faN \\
\texttt{\textbf{[P]}} Parametric RAG~\citep{su2025parametric} & \faN & \faH & \faN & \faN & \faH & \faN & \faN \\
\midrule
\makecell[l]{\texttt{\textbf{[M]}} \textbf{\nameAS [This work]}} & \faY & \faY & \faY & \faY & \faY & \faY & \faY  \\
\bottomrule
\end{tabular}
\vspaceSQ{-1.5em}
\end{table*}

\section{Related Work \& Advantages of \nameA}

\textbf{RAG Solutions.} We compare \nameAS to a large number of both traditional and modern {RAG solutions} in Table~\ref{tab:qualitative} in terms of design advantages (complexity analysis and empirical evaluation are in, respectively, Sections~\ref{sec:complexity} and~\ref{sec:eval}).
While prior RAG systems support retrieving multiple documents for a single query (e.g., RAG~\citep{lewis2020retrieval}, EMDR~\citep{singh2021end}), none of them generate \textit{multi-aspect} embeddings {per document}, and therefore do not offer multi-aspectuality. Similarly, although these systems employ Transformer architectures with MHA, they make no use of the MHA internal structure. Other methods such as Poly-encoder~\citep{humeau2019poly} and ColBERT~\citep{khattab2020colbert} do produce multiple embeddings per document, but for a different reason: they are based on models such as BERT, which inherently yields token-level embeddings. Thus, these models require multiple vectors per document simply to cover its content at the token level, not to represent distinct semantic aspects. In contrast, \nameAS leverages the powerful internal structure of modern decoder-based LLMs, where a small number of vectors derived from MHA (or even a single vector from the final decoder block, as in standard RAG) suffices to represent the semantics of an entire document or a chunk, as indicated by recent work~\citep{lee2024nv, besta2024checkembed}.

Building on this foundation, \nameAS achieves further practical advantages. It introduces \textit{scalable preprocessing} since (1) global embeddings are computed in a single forward pass per document and (2) scoring of heads is straightforwardly parallelizable. At inference time, it incurs \textit{no additional cost} relative to standard RAG -- in contrast to token-level approaches, which require computing numerous pairwise token similarities during retrieval, our method only compares compact, chunk-level vectors. \textit{Storage overhead is also minimal}: the MHA-derived embeddings match the dimensionality of standard last-layer embeddings (detailed time and storage complexity analyses are in Section~\ref{sec:complexity}). Finally, \nameAS requires \textit{no training} and is \textit{highly versatile}: it can be applied to any Transformer model with MHA, and is easily extensible to other modalities such as vision and graph data, where Transformer architectures are now common.

\textbf{Sparse retrievers} (e.g., BM25) rely on inverted indexes and lexical matching~\citep{robertson2009probabilistic}, while \textbf{learned sparse} retrievers (e.g., SPLADE) train models that output sparse lexical expansion vectors~\citep{formal2021splade}. \textbf{Hybrid} pipelines combine sparse and dense candidate sets via rank fusion (e.g., RRF)~\citep{cormack2009reciprocal}. \nameA\ is orthogonal to these approaches: it targets the \emph{dense retriever representation} and provides a training-free, lightweight multi-aspect embedding scheme that can serve as the dense branch in hybrid retrieval.

\textbf{Reranking.} Retrieval can be enhanced by \textbf{reranking}~\citep{rosa2022defense, nogueira2020passage, nogueira2020document, li2021parade, gao2021rethink, MacAvaney_2019}, where an initial candidate set is re-ordered using more precise scoring. Cross-encoder rerankers jointly encode query--document pairs via cross-attention, improving accuracy but incurring \textit{pairwise} scoring cost~\citep{nogueira2020passage, sbert_retrieve_rerank}. \nameAS uses a lightweight reranking scheme by default, but it is reranker-agnostic and can be seamlessly combined with existing cross-encoders without changes to the retriever or index.

\textbf{Multi-Head Embeddings Outside RAG.} Several methods propose a new model variant or architectural modification to the Transformer in order to better exploit MHA~\citep{park2020mhsan, huang2019multi, wang2020multi, xue2023pit}. For example, MHSAN~\citep{park2020mhsan} introduces a novel visual-semantic embedding network that extracts multiple region- and phrase-sensitive features using MHA. \citet{wang2020multi} develop a speaker embedding network that explicitly enforces head-level diversity via contrastive learning across resolutions. \citet{xue2023pit} propose PIT, a Transformer variant that composes attention heads across layers to reduce redundancy and enable efficient inference. \citet{vashisht2025mage} introduce MAGE, a technique that mixes heads across models to improve generalization through stochastic head combinations.
{In contrast, \nameAS is the first method to harness embeddings from MHA in pre-trained, decoder-based embedding LLMs for the purpose of more effective RAG. Unlike the above works, \nameAS requires {no architectural modifications, no training, and no custom modules}, being fully plug-and-play.}

\section{Conclusion}

\iftr
Retrieval-Augmented Generation (RAG) is crucial for democratizing access to accurate and relevant outputs from large language models (LLMs). However, enhancing the precision of RAG is non-trivial, especially given the challenges posed by queries requiring the retrieval of multiple documents with significantly different content. These complex multi-aspect queries are common across various domains, but existing RAG solutions struggle with them because the embeddings of the necessary documents can be far apart in the embedding space, complicating their retrieval.

To address this issue, we introduce \nameS (\nameA), a novel RAG scheme based on a simple yet powerful idea: leverage the activations from the multi-head attention (MHA) layer of decoder models instead of the traditional feed-forward layer. This approach is based on the insight, confirmed by an extensive literature survey, that different attention heads can capture distinct aspects of the data even without additional training. By using these diverse activations, \nameAS creates embeddings that better represent the multifaceted nature of data items and queries, thus enhancing the retrieval accuracy for complex, multi-aspect queries. The simplicity and versatility of this idea allow it to be seamlessly integrated into any modern RAG pipeline or data analytics framework, and to be applied to other classes of models beyond LLMs.

Our comprehensive evaluation methodology for multi-aspect queries, including novel metrics, synthetic datasets, and real-world use cases, demonstrates \nameA's effectiveness. The results indicate a significant improvement in the relevance of retrieved documents, with up to 20\% better performance compared to modern RAG baselines. Moreover, harnessing \nameAS results in enhanced downstream LLM generation by providing richer and more factual LLM outputs. This validates \nameA's potential to handle the intricacies of multi-aspect queries effectively.

Our analysis of nearly 20 RAG baselines illustrates that \nameAS is both highly effective and efficient because -- unlike other RAG schemes -- it does not require additional LLM queries, multiple model instances, increased storage, additional training, or multiple inference passes over the embedding model. These advantages, combined with the enhanced retrieval accuracy and \nameA's scalability, make \nameAS a valuable design in the field of LLMs and RAG systems.
\fi

\maciej{Analyze head specialization. See the commented Appendix. If not really the case, look for other explanations on why this works better for some cases}

\maciej{Address stuff from ICLR 2026 \url{https://openreview.net/forum?id=kppkHCZ14y}}

\maciej{Try more baselines} 

\maciej{Retry existing datasets}

\ifnbld
\section*{Acknowledgements}
We thank Hussein Harake, Colin McMurtrie, Mark Klein, Angelo Mangili, and the whole CSCS team granting access to the Ault, Daint and Alps machines, and for their excellent technical support.
We thank Timo Schneider for help with infrastructure at SPCL.
This project received funding from the European Research Council (Project PSAP, No.~101002047), and the European High-Performance Computing Joint Undertaking (JU) under grant agreement No.~955513 (MAELSTROM). This project was supported by the ETH Future Computing Laboratory (EFCL), financed by a donation from Huawei Technologies. This project received funding from the European Union’s HE research and innovation programme under the grant agreement No. 101070141 (Project GLACIATION).
We gratefully acknowledge Polish high-performance computing infrastructure PLGrid (HPC Center: ACK Cyfronet AGH) for providing computer facilities and support within computational grant no.~PLG/2024/017103.
We thank Kurt Krieg for useful insights into the nature of processing plant accidents.
\fi

\iftr
\phantomsection
\addcontentsline{toc}{section}{References}
\fi

\bibliographystyle{icml2026}
\bibliography{references.complete}

\newpage
\appendix

\section{Review of Multi-Aspectuality in Industry}
\label{sec:appendix-aspects}

Real-world decision-making and analysis tasks often require a \textit{holistic integration of multiple, semantically distinct information sources}. Literature from diverse domains---from safety investigations to diagnostics and enterprise analytics---consistently emphasizes that no single data stream suffices for complex tasks. Instead, success comes from combining heterogeneous elements (design details, human factors, environmental context, etc.) into a unified understanding. Below we highlight several examples that support this multi-aspect reasoning framework.

In \textbf{safety-critical environments}, accident and incident investigations rely on aggregating information from disparate sources. For example, a U.S.~Federal Railroad Administration report stresses that collecting multiple sources of information is essential for effective event reconstruction~\citep{fra2003human}. The {Columbia Accident Investigation Board} similarly combined telemetry, debris forensics, environmental data, and even autopsy reports to reconstruct the chain of events behind the Space Shuttle Columbia disaster~\citep{gehman2003columbia}. These investigations exemplify the need to integrate technical, procedural, and human-centered aspects of data to obtain actionable conclusions.

In \textbf{medicine}, particularly oncology, the concept of \textit{integrated diagnostics} exemplifies multi-aspect decision-making. Physicians routinely combine patient histories, radiological scans, lab values, pathology slides, and genetic tests to form a diagnosis. This is no longer optional: as modern datasets become more complex, integrating semantically distinct modalities has become necessary to reach accurate, personalized outcomes~\citep{messiou2023multimodal, kalia2013personalized}.

The same applies to \textbf{enterprise settings}. For example, airport operations management integrates foot traffic sensors, weather feeds, and gate schedules -- among others -- to optimize personnel allocation and prevent congestion~\citep{zografos2013decision}. Similarly, in legal and financial firms, cross-silo systems integrate internal metrics (e.g., billing and staffing) with external sources (e.g., news, contracts, social media) to guide decision-making and strategic planning~\citep{opentext}. These examples show that modern organizational workflows demand the integration of multiple semantically distinct data sources.

In summary, across domains like industrial safety, healthcare, and business intelligence, it is widely recognized that \textit{multi-aspectuality}---combining diverse, independently relevant information fragments---is essential for accurate and effective decision-making~\citep{fra2003human, gehman2003columbia, reason2006revisiting, kalia2013personalized, packham2017columbia, messiou2023multimodal, multer2013developing, coury2010transportation, harle2017investigation, o2000wheel, gordon2005designing, bridger2021guide, opentext}. In all these cases, the embeddings of documents from such divergent subdomains would be far away from one another in the embedding space when using standard RAG pipelines (as also confirmed by our own datasets in legal and plant accident use cases, see Section~\ref{sec:eval}), underlying the relevance of \nameA.

\section{Mathematical Formulation: Additional Details}
\label{sec:appendix-maths}

We omit, for clarity, unnecessary details such as layer normalizations.
The output of attention head $h$ for the $i$th token $x_i$ is defined as follows~\citep{vaswani2017attention}:

\begin{gather}
\text{head}^h (\mathbf{x}_i) = \sum_j w_{ij} \mathbf{v}_j^h, \quad \text{where} \nonumber \\
\vspaceSQ{-0.5em}
w_{ij} = \text{softmax}\left(\left(\mathbf{q}_i^h \right)^T \mathbf{k}_j^h\right), \nonumber \\
\vspaceSQ{-0.5em}
\mathbf{q}_i^h = \mathbf{W}_q^h \mathbf{x}_i, \quad \mathbf{k}_j^h = \mathbf{W}_k^h \mathbf{x}_j, \quad \mathbf{v}_j^h = \mathbf{W}_v^h \mathbf{x}_j, \nonumber
\end{gather}

\noindent
where $\mathbf{W}_q^h, \mathbf{W}_k^h, \mathbf{W}_v^h$ are, respectively, learnable query, key, and value projections associated with head $h$, and $\mathbf{x}_j$ is the vector embedding of the $j$th token $x_j$. These outputs get combined to form the output of the $i$th multi-head attention block as follows:

\begin{equation}
\text{MHA}(\mathbf{x}_i) = \mathbf{W}_o\  \text{concat}(\text{head}^1(\mathbf{x}_i), \dots, \text{head}^h (\mathbf{x}_i))^T, \nonumber
\end{equation}

\noindent
where matrix $\mathbf{W}_o$ is the linear layer that combines the outcomes of all attention heads.

\subsection{Standard RAG Embedding}

In standard RAG, a single embedding vector $\mathbf{e}_\text{std} \in \mathbb{R}^d$ is computed for a chunk by extracting the decoder output for the final token $x_n$ after the final feed-forward layer:

\begin{equation}
    \mathbf{e}_\text{std} = \text{FFN}\left( \text{MHA}(\mathbf{x}_n) \right) \nonumber
\end{equation}

\subsection{Multi-Head RAG Embedding}

Instead of compressing all head outputs into a single embedding, MRAG leverages the individual output of each head on the final token $x_n$:

\begin{equation}
    \mathcal{S} = \{ \mathbf{e}_k = \text{head}^k(\mathbf{x}_n) \in \mathbb{R}^{d/h} \mid k = 1, \dots, h \} \nonumber
\end{equation}

This results in $h$ head-wise embeddings per chunk, capturing diverse semantic aspects. Crucially, this design avoids any increase in memory or compute during inference, as head-level vectors are computed as part of the standard MHA process.

\section{System Design \& Implementation: Additional Details}
\label{sec:app:system-dets}

We provide additional details on system design and implementation.

\subsection{Ranking Strategy Details}
\label{sec:app:system-dets:scoring}

The scoring of embedding spaces is detailed in Algorithm~\ref{algo:indexing}.

%
      \begin{algorithm}[H]
      \footnotesize
        \caption{Importance scores for heads.}
        \label{algo:indexing}
        \begin{algorithmic}
          \FOR{each head $h_i$}
            \STATE{$a_i, b_i, count\_a_i, count\_b_i \leftarrow 0$}
            \FOR{each embedding $e_{ij}$ in $h_i$}
                \STATE{$a_i \leftarrow a_i + ||e_{ij}||$}
                \STATE{$count\_a_i \leftarrow count\_a_i + 1$}
                \FOR{subset of $m$ embeddings $e_{ih}$ sampled uniformly at random} 
                    \STATE{$b_i \leftarrow b_i + \text{cosine-distance}(e_{ij}, e_{ih})$}
                    \STATE{$count\_b_i \leftarrow count\_b_i + 1$}
                \ENDFOR
            \ENDFOR
            \STATE{$a_i \leftarrow a_i / count\_a_i$; $b_i \leftarrow b_i / count\_b_i$}
            \STATE{$s_i \leftarrow a_i \cdot b_i$}
          \ENDFOR
        \end{algorithmic}
       \end{algorithm}

\subsection{Ranking Strategy Details}
\label{sec:app:system-dets:ranking}

The voting strategy used by \nameAS in its reranker is pictured in Algorithm~\ref{algo:voting}.

      \begin{algorithm}[H]
        \caption{Voting strategy.}
        \label{algo:voting}
        \begin{algorithmic}
          \STATE{$l\leftarrow[]$}
          \FOR{each head $h_i$ and its score $s_i$}
            \STATE{find best matching $c$ text chunks}
            \FOR{each chunk $d_{i,p}$ with index $p$ in top $c$}
                \STATE{$w_{i,p} \leftarrow s_i\cdot 2^{-p}$}
                \STATE{add tuple $(d_{i,p}, w_{i,p})$ to $l$}
            \ENDFOR
          \ENDFOR
        \STATE{sort $l$ using weights $w_{i,p}$}
        \STATE{return top $k$ elements of $l$}
        \end{algorithmic}
      \end{algorithm}

\subsection{Integration with Data Stores}
\label{sec:app:system-dets:integration}

\nameAS can be seamlessly used with different classes of data stores \ipartc~and nearest neighbor (NN) search approaches. It can be combined with both exact and approximate NN to find the matching (embedding, chunk)-pairs.
These two parts of the broader RAG processing pipeline are orthogonal to \nameA.

\newpage
\onecolumn
\section{Specification of Prompts}
\label{sec:app:prompts}

\definecolor{darkgrey}{HTML}{4A4A4A}
\definecolor{lightgrey}{HTML}{CCCCCC}

\newtcolorbox{prompt}[2][]{%
  enhanced,
  breakable,
  colframe=darkgrey,
  colback=white,
  title style={fill=darkgrey},
  coltitle=white,
  fonttitle=\bfseries,
  title=#2,
  fontupper=\ttfamily\small,
  #1
}


\subsection{Prompt Template for Reader}

\begin{prompt}{Prompt Template for Reader}

    You are presented with a series of articles, each potentially addressing different aspects of a topic.\\

	1. \textless{}article title\textgreater{}\\
	{[\textless{}metadata\textgreater{}]} \\
	\textless{}body\textgreater{} \\
    --------- \\

    2. \textless{}article title\textgreater{}\\
	{[\textless{}metadata\textgreater{}]} \\
	\textless{}body\textgreater{} \\
    --------- \\

	\textless{}...\textgreater{} \\
	--------- \\

    Task: Carefully analyze the articles. When formulating your response to the question below, identify the relevant aspects or claims made within each article. Construct your answer by comparing, contrasting, or synthesizing these points in a coherent and logically structured manner. Your response should be supported by specific references to the content of the articles. Where applicable, acknowledge differences in perspectives, data, or assumptions across the sources. Aim for clarity, precision, and concise reasoning grounded in the evidence provided.\\

    Please answer the following question: \textless{}query text\textgreater{}\\

\end{prompt}

\subsection{Prompt Template for the Synthetic Dataset Generation}

\begin{prompt}{Prompt Template for Query Generation}
    Please create a story about the attached \textless{}number of articles\textgreater{} articles on the topics \textless{}list of titles\textgreater{}. \\

	It is very important that each of the attached articles is relevant to the story, in a way that references the content of the article, not just its title. But please also mention each title at least once. Please make sure that all of the attached articles are relevant to your story, and that each article is referenced in at least two sentences! They do not necessarily have to be referenced in the same order, but make sure no article is forgotten. \\

	Important: Output only the story, no additional text. And do not use bullet points, or paragraphs. \\

	Articles: \\
	--------- \\
	Article \textless{}title\textgreater{}: \\
	\textless{}body\textgreater{} \\

	\textless{}...\textgreater{} \\
	--------- \\
	Again, make sure that you reference all the following topics in your story: \textless{}list of titles\textgreater{} \\

\end{prompt}

\twocolumn

\section{Multi-Aspectuality with Attention Heads without Additional Training}
\label{sec:app:head_analysis}

In \nameA, we extract embeddings from the hidden representations immediately after the attention block in the last decoder layer, avoiding any fine-tuning. This decision is based on an existing hypothesis that attention heads in Transformer models naturally differentiate during training, each attending to distinct aspects of the input data distribution.

\subsection{Literature Survey}
\label{sec:app:heads-survey}

This hypothesis has been substantiated for various model families. \citet{wang2024differentiation} introduce the \emph{Local Learning Coefficient} (LLC), a measure of training dynamics at the head level. They show that attention heads begin with similar behavior but quickly diverge into functionally distinct clusters, each specializing in different patterns, ranging from local structure to multigram token groups. \citet{olsson2022incontext} identified {``induction heads''} in GPT-style models: specific heads that attend to earlier repeated sequences (e.g., in patterns like ``X ... Y ... X''), enabling the model to learn simple in-context repetition \textit{without additional supervision}. Further studies observed heads that consistently attend to names, suppress repetition, or shift attention predictably to structurally aligned tokens~\citep{mcdougall2024copy, gould2024successor}.

Research in BERT-style models further supports these trends. \citet{clark2019bert} demonstrate that different heads attend to direct objects, nominal modifiers, or punctuation tokens. \citet{kovaleva2019revealing} identify broad attention patterns like ``vertical'' heads (focusing on special tokens) and ``diagonal'' heads (attending to adjacent words). \citet{htut2019attention} show that many heads correlate with syntactic dependency arcs.

Notably, while many heads specialize in meaningful ways, others appear to contribute little to model performance. \citet{voita2019analyzing} and \citet{michel2019sixteen} show that a large fraction of heads can be pruned without substantial performance loss, suggesting that specialization tends to concentrate in a smaller subset of effective heads.

\subsection{Analysis of Multi-Head Patterns}
\label{sec:app:heads-analysis}

We investigated the attention heads of two models in detail: LLaMA-2 7B and
SFR-Embedding-Mistral. We selected these two models for a detailed investigation
because the former represents models that are not fine-tuned for text
embeddings, while the latter is specifically the text embedding model that we
used for our experiments. For each model, we looked specifically at the
attention scores within each attention head, i.e., how much attention each head
pays to each input token during the inference. Knowing the semantics of the
input tokens enables then deriving certain conclusions about multi-aspectuality
and attention heads.

\begin{figure*}[t]
  \centering
  \begin{subfigure}[t]{0.24\textwidth}
    \includegraphics[width=\textwidth]{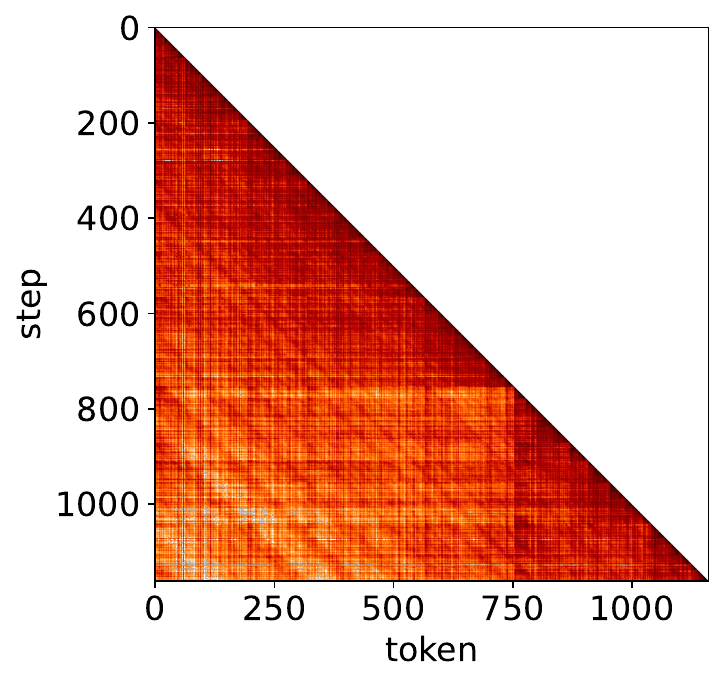}
    \caption{Head 4 LLaMA-2}
    \label{fig:headmap_head_4_llama}
  \end{subfigure}
  \hfill
  \begin{subfigure}[t]{0.24\textwidth}
    \includegraphics[width=\textwidth]{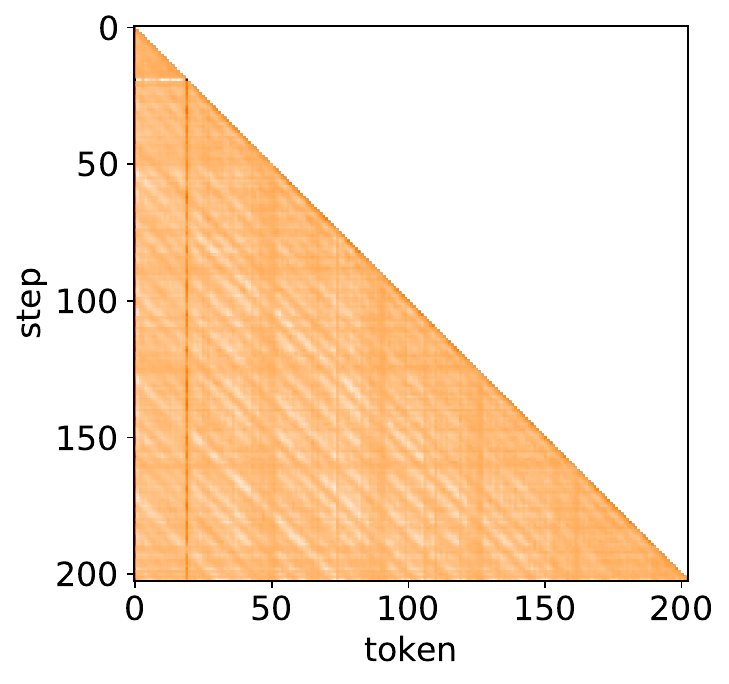}
    \caption{Head 8 SFR}
    \label{fig:headmap_head_8_sfr}
  \end{subfigure}
  \hfill
  \begin{subfigure}[t]{0.24\textwidth}
    \includegraphics[width=\textwidth]{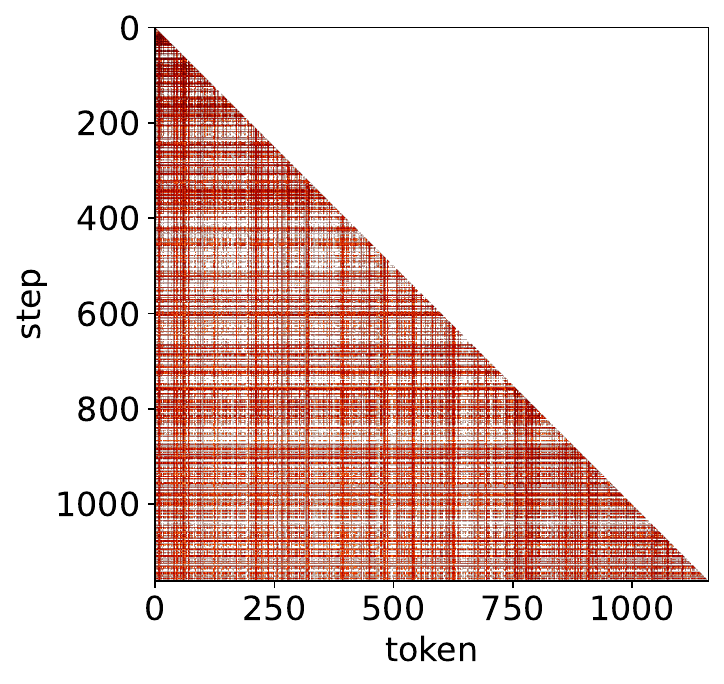}
    \caption{Head 22 LLaMA-2}
    \label{fig:headmap_head_22_llama}
  \end{subfigure}
  \hfill
  \begin{subfigure}[t]{0.24\textwidth}
    \includegraphics[width=\textwidth]{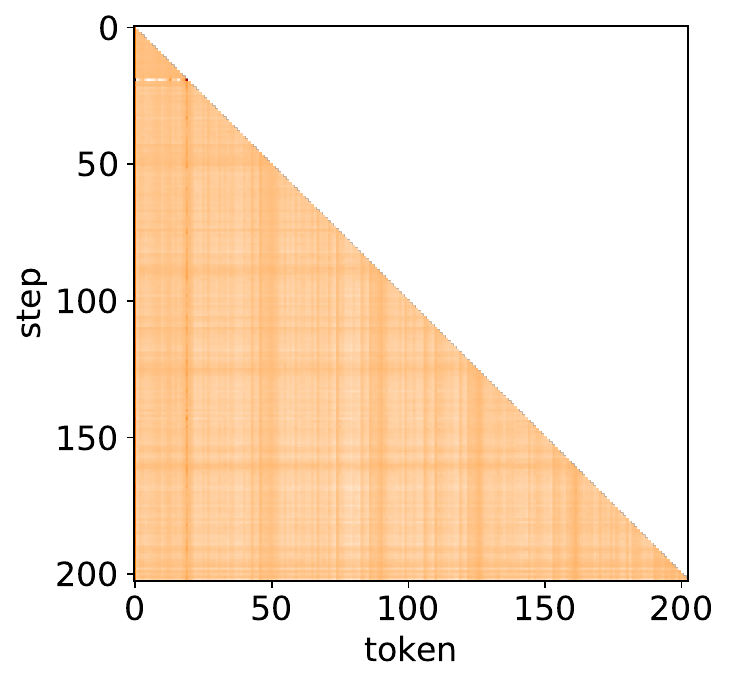}
    \caption{Head 21 SFR}
    \label{fig:headmap_head_21_sfr}
  \end{subfigure}
  \caption{Heatmap plots for selected attention heads of the LLaMA-2 7B and SFR-Embedding-Mistral models.}
  \label{fig:heatmap}
\end{figure*}

We plot selected results in Figure~\ref{fig:heatmap}. Each heatmap shows the
dot-product between key- and value-projections inside a given specified
attention head, where line $i$ of a heatmap for attention head h indicates the
dot-products between the query-projection of token $x_i$ and the key-projections of
all previous tokens $j < i$ (both models use causal attention).

For both models, we found that the attention patterns vary significantly
between the different attention heads. Still, we encountered two distinct
patterns. First, the diagonal lines in Figures~\ref{fig:headmap_head_4_llama}
and~\ref{fig:headmap_head_8_sfr} indicate that, when processing a certain input
token $x$, elevated attention is paid to some tokens that came a constant number
of steps before $x$. We postulate that this pattern is likely beneficial to
understanding the overall rhythm of a natural language, allowing the model to
better identify which words are semantically connected, and which parts of the
input text refer to each other. Second, horizontal and vertical lines in
Figures~\ref{fig:headmap_head_22_llama} and~\ref{fig:headmap_head_21_sfr} show
that these heads learnt to pay attention to specific tokens, regardless of how
far apart they are within the input sequence. An intuitive justification for
such patterns is the focus on certain semantic aspects of the input sequence.

\begin{figure*}[t]
  \centering
  \begin{subfigure}[t]{0.48\textwidth}
    \includegraphics[width=\textwidth]{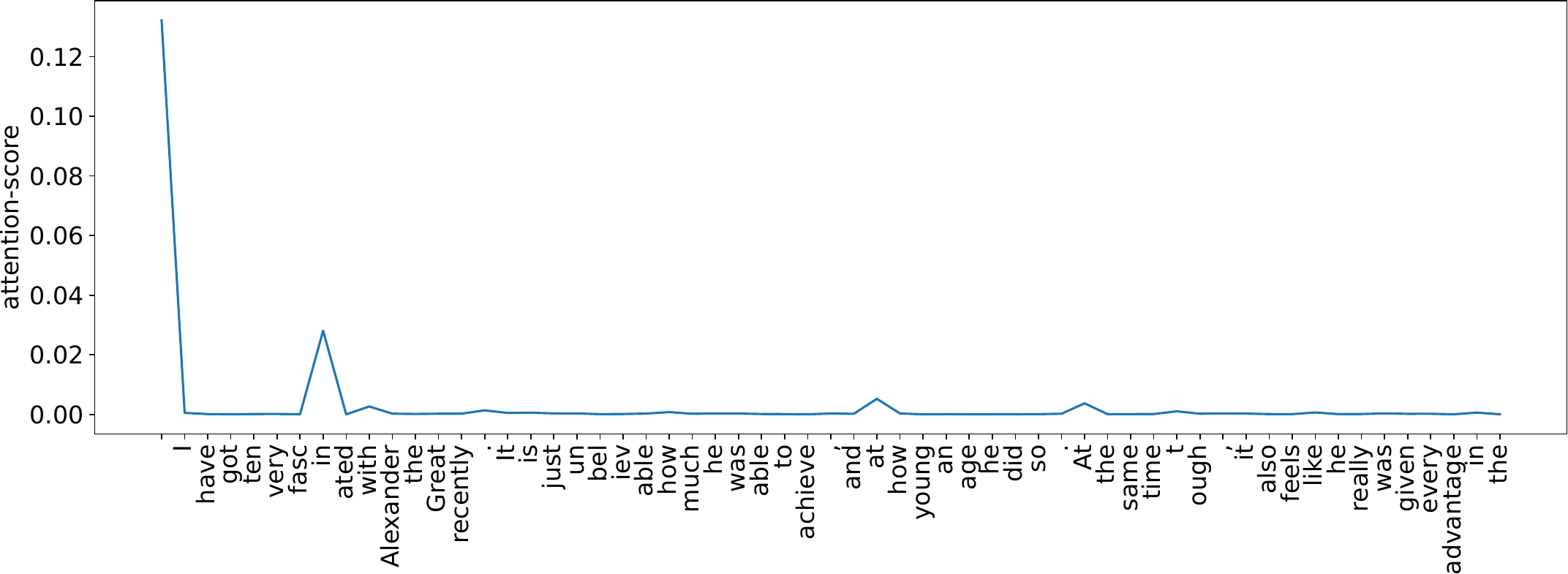}
    \caption{Attention Head 0 LLaMA-2 7B}
  \end{subfigure}
  \hfill
  \begin{subfigure}[t]{0.48\textwidth}
    \includegraphics[width=\textwidth]{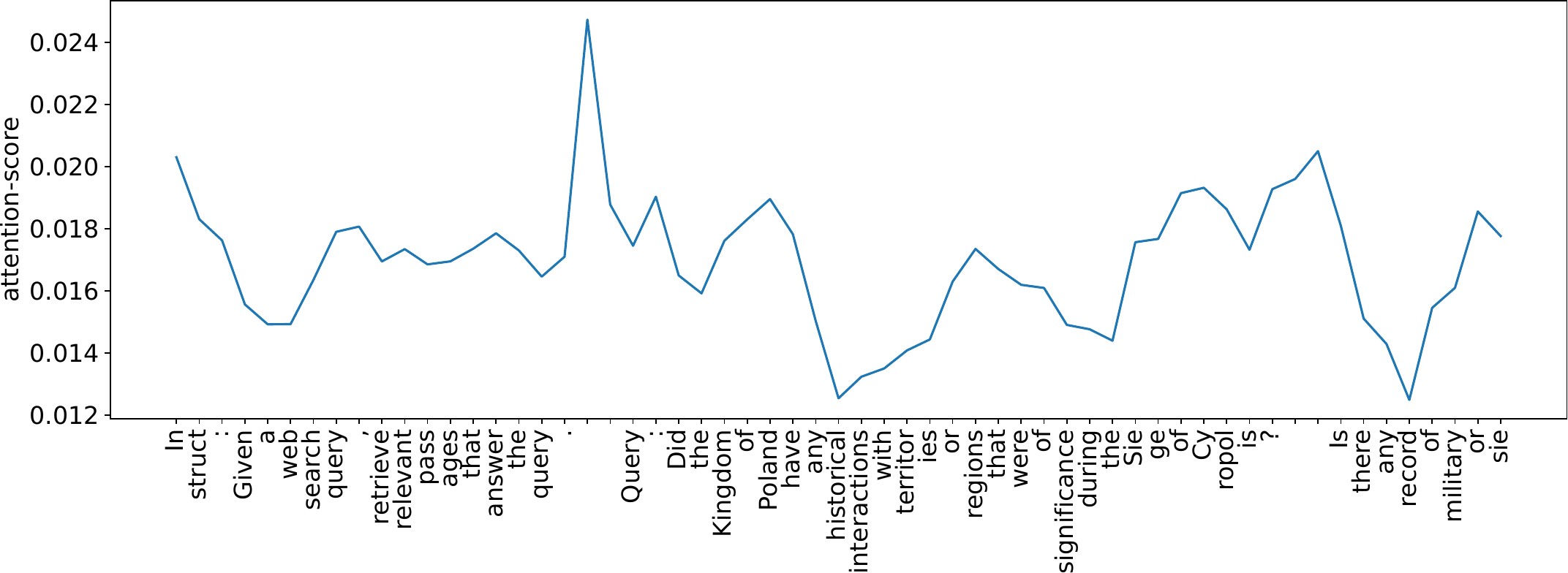}
    \caption{Attention Head 8 SFR}
  \end{subfigure}
  \begin{subfigure}[t]{0.48\textwidth}
    \includegraphics[width=\textwidth]{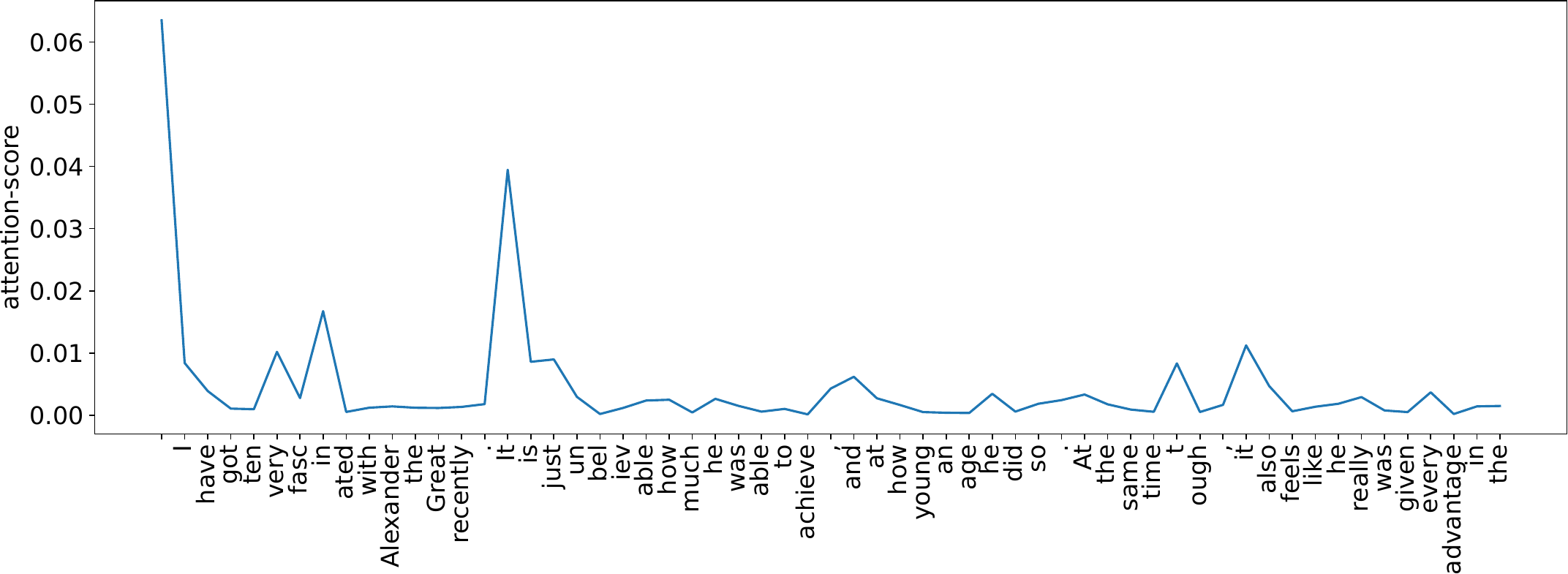}
    \caption{Attention Head 14 LLaMA-2 7B}
  \end{subfigure}
  \hfill
  \begin{subfigure}[t]{0.48\textwidth}
    \includegraphics[width=\textwidth]{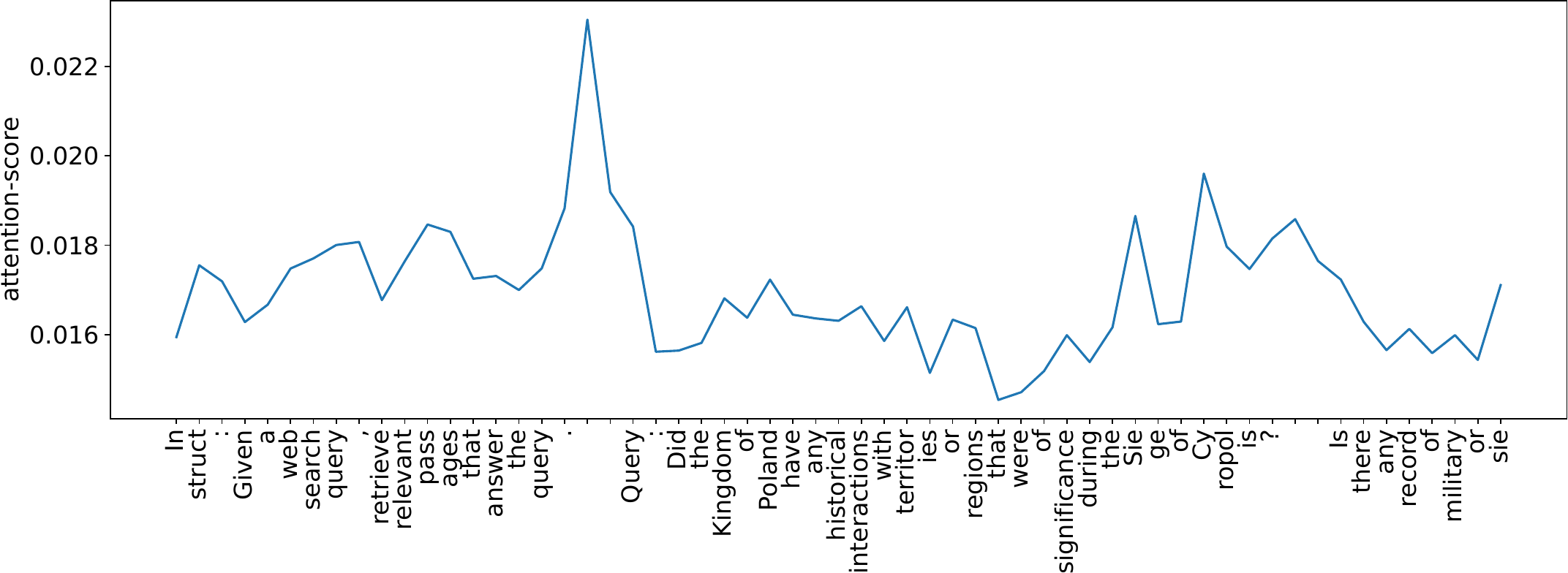}
    \caption{Attention Head 21 SFR}
  \end{subfigure}
  \caption{Attention scores for selected attention heads of the LLaMA-2 7B and SFR-Embedding-Mistral models.}
  \label{fig:attention_scores_heads}
\end{figure*}

\begin{figure*}[t]
  \centering
  \includegraphics[width=0.8\textwidth]{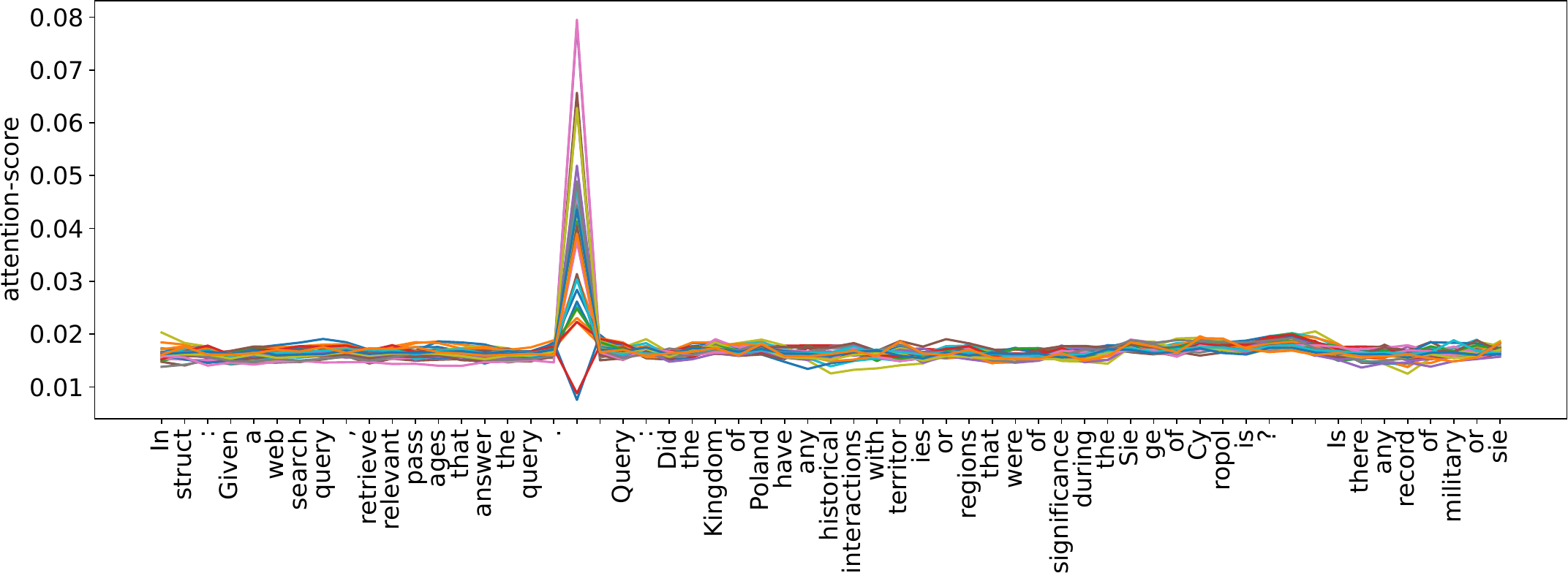}
  \caption{Attention scores for all attention heads of the SFR-Embedding-Mistral model.}
  \label{fig:attention_scores}
\end{figure*}

We also detail attention scores (after applying softmax) of selected heads in
Figures~\ref{fig:attention_scores_heads} and~\ref{fig:attention_scores}, when
the model is processing the last token of its input. We see that some tokens
gather a lot of attention from most heads, yet there is always a plethora of
passages which are attended differently by any two attention heads. An
interesting pattern we encountered was that for the SFR-Embedding-Mistral model
(see Figure~\ref{fig:attention_scores}), all heads' attention spiked
significantly on the first line-break in the input sequence - either positively
or negatively. We conjecture that this is a consequence of how the embedding
model was fine-tuned and its intended usage pattern: embedding queries are
usually prepended with a retrieval instruction, which is terminated by a
line-break. The model likely learnt to summarize the necessary information
about this instruction inside the terminating line-break.

\section{Complexity Analysis: Additional Details}
\label{sec:app:complexity}

We now discuss additional details for our complexity analysis of the RAG schemes.

In Poly-encoders~\citep{humeau2019poly} retrieval, we use $s(n)$ to represent additional attention evaluations needed to prepare the $m$ code embeddings and their final context embeddings with all $n$ documents.

In \citet{lewis2020retrieval} retrieval, the work and depth are increased by the additional generator model evaluation.

In ColBERT~\citep{khattab2020colbert} retrieval, the work is increased linearly and depth logarithmically with the average query and document size due to the \verb|maxsim| evaluations. For storage, the overhead scales linearly with the average size of the document, due to per-token embeddings.

In EMDR~\citep{singh2021end} retrieval, the work is increased $k+1$ times due to \verb|T5| encoders run on the top $k$ documents. This corresponds to a $D_m$ depth increase as the evaluations can be parallelized.

In Self-RAG~\citep{asai2023selfrag}, once the model decides to conduct retrieval using the \verb|retrieval| token, it embeds the query, runs nearest $k$ search, and evaluates the model twice for each best-$k$ document to predict the \verb|issup|, \verb|isrel|, \verb|isuse| labels. The depth increases only by $D_m$ as the evaluations can be parallelized.

In Chain-of-Note~\citep{yu2023chain}, during retrieval the $k$ best documents are fetched and then each is summarized sequentially by generating notes, resulting in a $ks(n)$ work and depth increase, with $s(n)$ representing the cost to generate an average note.

In RAPTOR~\citep{sarthi2024raptor}, preprocessing involves creating a document tree, with leaves representing documents, and parents containing summaries of children, a cost we represent with $s(n)$ both for work and storage. Retrieval involves traversing the tree, which we encapsulate in a different $s(n)$.

RAGraph~\citep{jiang2024ragraph} preprocessing involves embedding the documents by a graph model and creating the toy graphs, which we account for using $W_e$ and $D_e$. The toy graphs also create additional space requirements, which we account for using $s(n)$. For retrieval, additional key information such as environment or position-aware codes is used, represented by another $s(n)$ in our notation.

In RQ-RAG~\citep{chan2024rq}, the retrieval may be decomposed into multiple separate queries that are generated by the model and then evaluated in sequence using standard techniques, resulting in increases in work and depth that we denote using $s(n)$.

In ThinkNote~\citep{xu2024active}, retrieval is conducted using standard techniques, followed by three separate model evaluations: the Knowledge Assimilation (KA), Self-Inquiry (Q), and Thought Accommodation (TA) agents. As TA is similar to the work of an LLM answering the query, we omit its cost, like in all other frameworks, resulting in triple the work. Q and KA can be evaluated in parallel, increasing the depth only twice.

HiQA~\citep{chen2024hiqa} combines multiple retrieval strategies, using vector similarity matching, elastic search with BM25, and keyword matching. We represent these with $s(n)$ for work. Note that as these are independent, the depth is not increased. For preprocessing, HiQA uses a Hierarchical Contextual Augmentor, which creates a data hierarchy and introduces an overhead we denote by $s(n)$. HiQA also stores more information alongside vectors, such as keywords, increasing requirements by $s(n)$.

GraphRAG~\citep{edge2024from} creates a knowledge graph during preprocessing, which we denote with $W_e$ and $D_e$. It then extracts communities and their summaries, which we account for using $s(n)$. These are stored and require additional space also denoted by a different $s(n)$. During retrieval, GraphRAG uses an LLM to rank all communities in parallel by how useful they are which we estimate using $s(n)$.

In Fusion RAG~\citep{rackauckas2024rag} retrieval, $k$ queries are generated and for each a standard RAG is evaluated, together with a reranking achieved by another model evaluation, which we estimate as $s(n)$.

In Meta-chunking~\citep{zhao2024meta}, the documents are preprocessed by splitting them into chunks based on perplexity or margin sampling. As the decision whether to split or combine sentences in a document is based on an LLM, the preprocessing requires $l_d$ evaluations with some $s(n)$ postprocessing. The storage requirements are also increased by a different $s(n)$ as documents are stored not as a single vector but multiple vectors based on chunks. For the same reason, retrieval requires more work, as the number of vectors is increased by $s(n)$.

In MoC~\citep{zhao2025moc}, chunks are created based on different granularities. Similarly to Meta-chunking, this means a larger number of vectors resulting in increased storage requirements, and retrieval work by $s(n)$. As MoC also includes routing and meta-chunkers, the preprocessing cost is increased by $s(n)$.

In Parametric RAG~\citep{su2025parametric}, documents are represented as deltas of parameters that can be applied to the model. During preprocessing, the model is fine-tuned on a given document, resulting in a considerable increase of $s(n)$ in work and depth. As parameters are considerably larger than the hidden dimension, storage is also increased by a different $s(n)$. During retrieval, standard RAG fetches the documents, and the original model needs to be updated with their parameters, increasing work and depth by $s(n)$.

SuperRAG~\citep{yang2025superrag} first embeds the documents in a knowledge graph, which we represent by $W_e$ and $D_e$, and then uses this knowledge graph in retrieval to index it using $W_i$ and $D_i$. These also include any reranking SuperRAG might do.

HiRAG~\citep{huang2025retrieval} creates a hierarchical knowledge graph used for indexing in retrieval. Similarly, to SuperRAG, this increases the preprocessing and retrieval costs. We include in these the additional description and report generation that HiRAG conducts.

\section{Full Specification of Benchmarking Multi-Aspectuality}
\label{sec:app:multi-aspect-data-and-metrics}

\begin{figure*}[t]
    \centering
        \includegraphics[width=1.0\linewidth]{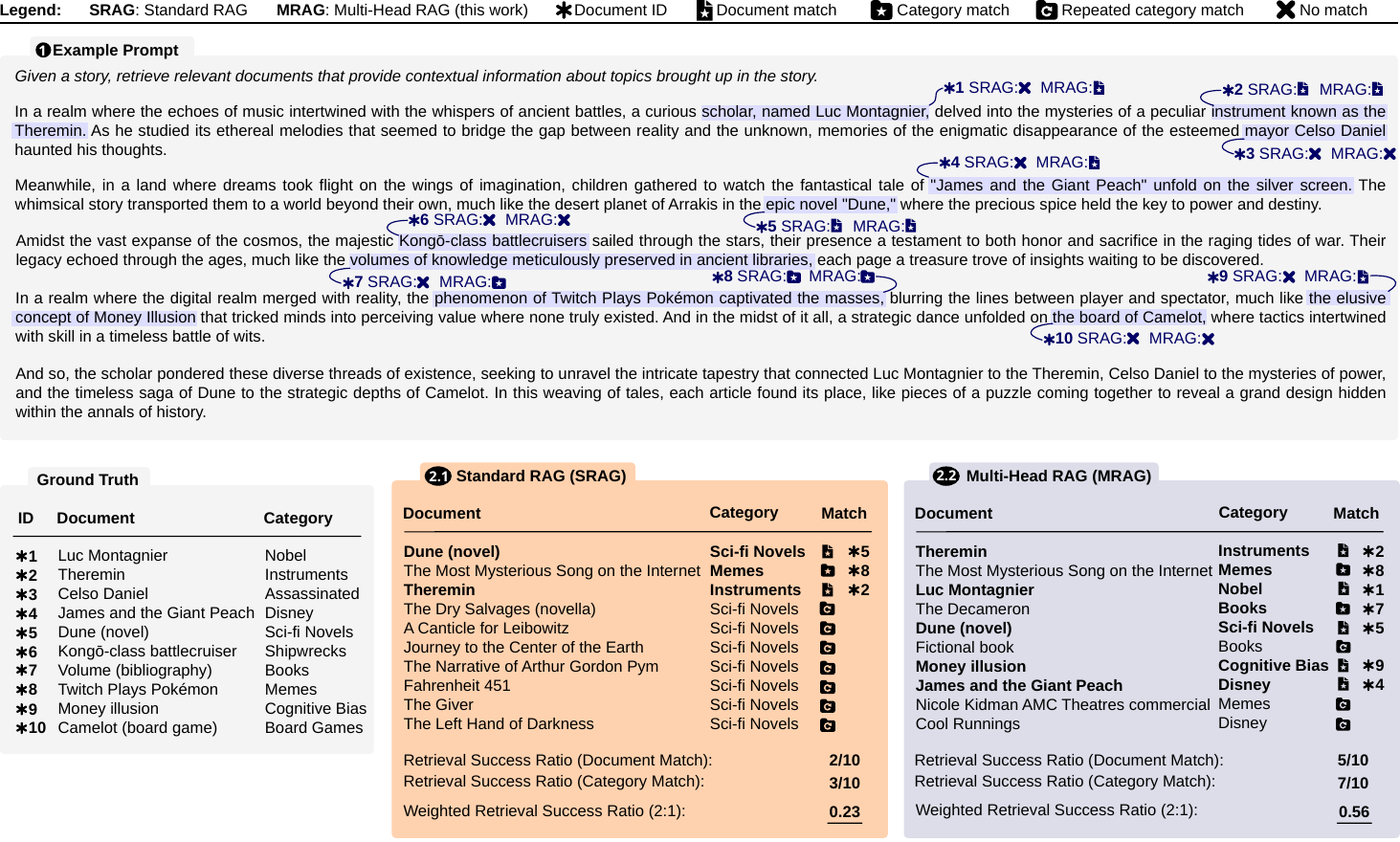}
    \caption{An example query used to evaluate different RAG strategies. We mention the documents to be fetched in the text and then assess the success ratio of different RAG strategies in finding these documents and their categories. We mark exact document matches \idmatch\thinspace, category matches \icmatch\thinspace, documents that match a category multiple times \icrematch\thinspace, and text segments with no matching document \inomatch\thinspace. Finally, we show the weighted success ratio for each strategy, taking a 2:1 weighting (prioritizing the exact article matches).}
    \label{fig:query_detail}
\end{figure*}

We provide more details on how to benchmark multi-aspectuality in RAG.
Figure~\ref{fig:query_detail} shows an example query and metrics usage. Each query requires retrieving a specific number of documents and the corresponding non-overlapping categories which define the {ground truth}. We fetch the top $k$ documents from a database, where $k$ is the ``total number of documents fetched for a tested RAG scheme'' (including potentially mismatches). Among these $k$ documents, we search for matches with the ground truth.


\subsection{Multi-Aspect Datasets}
\label{sec:multi-aspect-data}

We first select conceptually different categories of documents for a synthetic dataset. Here, we harness publicly available Wikipedia articles. In the dataset construction pipeline, the user selects a given number of categories (e.g., countries, board games, historical swords, shipwrecks, etc.) and then, for each category, they sample a specified number of documents. The first part of the document (overview) is used as a text chunk to be embedded. We enforce that each overview must have at least 800 characters, matching commonly used chunk sizes in RAG schemes. 
%
%
We also use multi-aspect \textbf{real-world inspired datasets} consisting of NDAs and reports describing industry accidents in chemical processing plants. We ensure the usefulness of these datasets by working directly with tech leaders from 3 corporations that rely on RAG in their in-house LLM-driven report generation and analytics frameworks. Example categories of the legal documents are legal areas (energy law, family law, criminal law, etc.) or document language style (aggressive, mild, neutral, etc.). Examples of accident causes are natural disasters, human mistakes, or lack of proper training. We fully release these datasets to propel RAG research. Details on all three datasets can be found in the Appendix~\ref{sec:datasets}.
In our evaluation, we use a total of 16,500 documents.

\subsection{Multi-Aspect Query Generation}
\label{sec:multi-aspect-queries}

We also require queries that touch upon a given \textit{number of $n$ aspects}. For example, a query with 10 aspects must contain a question about 10 different documents from 10 different categories. We create such queries by selecting $n$ categories, sampling a document from each selected category (ensuring there are no duplicates overall), and then generating a story that combines these documents, using an LLM (GPT-4o, GPT-4o mini).
We construct 160 queries with 1, 2, 3, 4, 5, 6, 10, 15, 20 and 25 aspects (1600 queries in total).
An example multi-aspect query sent to the LLM that requires retrieving 10 documents from 10 different categories, is pictured in the top part of Figure~\ref{fig:query_detail}.

\begin{table*}[t]
\footnotesize
  \centering
  \caption{Overview of the structure and the number of documents in the respective datasets.}
  \label{tab:dataset}
  \begin{tabular}{lcccc}
  \toprule
  \textbf{dataset} & \textbf{\#categories} & \textbf{\#topics} & \textbf{\#documents} & \textbf{total \#documents} \\
  \midrule
  Wikipedia & 80 & \multicolumn{2}{c}{50 documents per category} & {4000} \\
  Legal Documents & 25 & 25 per category & 10 per topic & {6250} \\
  Accident Reports & 25 & 25 per category & 10 per topic & {6250} \\
  \bottomrule
  \end{tabular}
\end{table*}

\subsection{Metrics}
\label{sec:multi-aspect-metrics}

We also design novel metrics to assess how well a given RAG scheme supports multi-aspectuality.
For a query $Q$, a used reranker scheme $S$ (detailed in Section~\ref{sec:query-exec}), and $n$ documents from $n$ categories to retrieve, $Q_{rel}$ denotes the \textit{ideal} set of documents that should be retrieved for $Q$. Then, $S(Q, n)$ is the set of the \textit{actually} retrieved documents. We define the \textit{Retrieval Success Ratio} as $\Xi(Q, n) = \frac{|S(Q, n) \cap Q_{rel}|}{|Q_{rel}|}$, i.e., the ratio of successfully retrieved relevant documents.
%
%
%
%
%
Moreover, there is a case when a RAG scheme does not retrieve the \textit{exact} desired document, but it still retrieves successfully \textit{some other document} from \textit{the same} category. While less desired, it still increases chances for a more accurate LLM answer following the retrieval. For example, when asking the LLM to determine the cause of an industry accident, fetching the documents in the same category as the accident being queried about, improves the chances for the LLM to give a more relevant answer. To consider such cases, we use another measure, the \textbf{Category Retrieval Success Ratio} or $\Xi_c$. It has the same form as $\Xi(Q, n)$ above, with one difference: $S(Q, n)$ is now the set of all the retrieved documents that belong to categories of the ideal desired documents. 
%
%
Finally, to combine these two metrics, we use the \textbf{Weighted Retrieval Success Ratio} $\Xi_w$ as $\Xi_w=\frac{w\cdot\Xi+\Xi_c}{w+1}$.
%
%
%
%
%

%
An example of using these metrics to assess how well \nameAS and Standard RAG capture multi-aspectuality is pictured in the bottom part of Figure~\ref{fig:query_detail}.

\section{Evaluation Setup: Additional Details}
\label{sec:app:setup}




\subsection{Compute Resources}
\label{sec:compute_resources}

Our experiments were executed with compute nodes containing 4$\times$ NVIDIA GH200 and a total memory of 800 GB. In general one GPU with at least 40GB of memory should suffice. We used at most 50GB of storage and the OpenAI API as an external resource. The full experiments took at most three hours of GPU time and the cost for the OpenAI API were at most \$15. We carried out additional experiments, which amounted to around 20 hours of GPU time and cost of \$25 for the OpenAI API. Additional evaluation was executed with a mix of compute resources including NVIDIA A100 and V100 GPUs.

\subsection{Dataset Details}
\label{sec:datasets}

We provide an overview over the used datasets in Table~\ref{tab:dataset}.

\section{Evaluation: Additional Results}
\label{sec:app:eval}

We provide additional empirical evaluation results.

\subsection{Harnessing Different Decoder Blocks}

\begin{figure}[t]
    \centering
        \includegraphics[width=0.98\columnwidth]{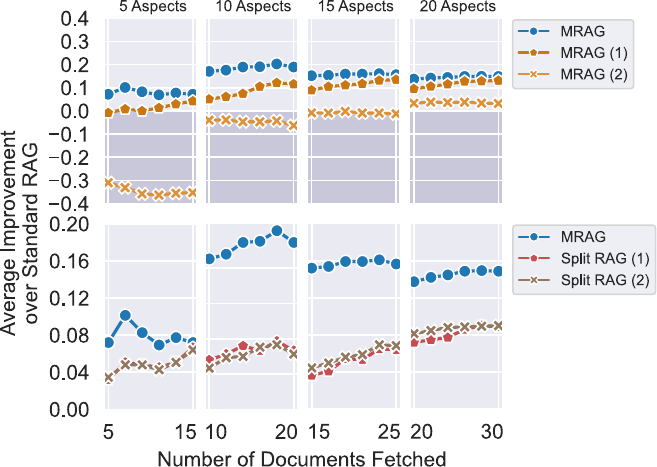}
    \caption{Evaluation of different voting strategies for \nameAS and Split RAG.}
    \label{fig:plot_additional}
\end{figure}

We analyze the impact of using embeddings from \textbf{different decoder blocks} for \nameAS (instead of the last one).
Here, we consider taking multi-aspect embeddings from three different layers of the embedding model: after the first multi-head attention block, after multi-head attention block~16 (in the middle of the decoder architecture), and the final multi-head attention. We discover that the last multi-head attention performs the best when compared with the Standard RAG.

\subsection{Analyzing Different Voting Strategies}

We also illustrate selected representative data from a long investigation into two \textbf{additional voting strategies} for \nameA. We compare \textbf{\nameAS (1)} where only the exponential lowering of significance of selected chunks is applied ($w_{i,p} = 2^{-p}$), and \textbf{\nameAS (2)} which assigns the weight for each text chunk based on the distance between the particular text chunk ($d_{i,p}$) and the query ($q$) ($w_i = \frac{1}{distance(d_{i,p}, q)})$. Figure~\ref{fig:plot_additional} (top part) shows that these voting strategies perform worse on average than our selected strategy for \nameA, justifying its design and selection (described in Section~\ref{sec:query-exec}).

We also consider two voting strategies for Split RAG, to further deepen the empirical evaluation. \textbf{Split (1)} only uses the exponential lowering of significance ($w_{i,p} = 2^{-p}$) and \textbf{Split (2)} which uses the same strategy as \nameAS ($w_{i,p} = s_i \cdot 2^{-p}$). Figure~\ref{fig:plot_additional} (bottom part) shows that these voting strategies are on par with each other while being worse than \nameA, further showcasing the advantages of \nameA.

\subsection{Analyzing Preprocessing Overhead}


One-time head importance scoring in \nameAS introduces minimal preprocessing overhead on top of the standard embedding scheme. The scoring consists of computing: (i) average L2 norms per embedding space, and (ii) average pairwise cosine distances among embeddings within each space.
To assess practical overhead, we analyze six datasets using the SFR-Embedding-Model. The additional time required to compute importance scores is measured as a percentage of the original embedding time. For example, on SciFact~\cite{wadden2020fact}, the overhead was just 2.7\%, and on NFCorpus~\cite{boteva2016fulltext}, only 1.75\%. Even for moderate-scale corpora such as ArguAna~\cite{wachsmuth2018retrieval} (310 seconds total encoding time), the overhead remained under 5\%. These results assume full pairwise distance computation across all chunks; in practice, the harnessed sampling makes them even lower.

\end{document}